\documentclass[conference]{IEEEtran}
\IEEEoverridecommandlockouts
\usepackage{cite}
\usepackage{amsmath,amssymb,amsfonts}
\usepackage{algorithmic, algorithm}
\usepackage{graphicx}
\usepackage{textcomp}
\usepackage{xcolor}
\usepackage{todonotes}

\usepackage{subcaption}
\usepackage{textcomp}
\usepackage{transparent}
\usepackage{makecell}
\usepackage{multirow}
\usepackage{color, colortbl}

\usepackage{array}
\newcolumntype{x}[1]{>{\centering\arraybackslash}p{#1}}
\usepackage[
    colorlinks=true,
    allcolors=black,
    pdftitle={XPROA-B}
]{hyperref}

\def\BibTeX{{\rm B\kern-.05em{\sc i\kern-.025em b}\kern-.08em
    T\kern-.1667em\lower.7ex\hbox{E}\kern-.125emX}}

\DeclareMathOperator*{\argmax}{arg\,max}

\makeatletter
\newcommand{\linebreakand}{
  \end{@IEEEauthorhalign}
  \hfill\mbox{}\par
  \mbox{}\hfill\begin{@IEEEauthorhalign}
}
\makeatother

\usepackage{xpatch}
\makeatletter
\xpatchcmd{\@thm}{\thm@headpunct{.}}{\thm@headpunct{}}{}{}
\makeatother

\newcommand{\hlBlue}[2][0.1]{{\transparent{#1}\colorbox{blue}{\transparent{1}#2}}}
\newcommand{\hlRed}[2][0.1]{{\transparent{#1}\colorbox{red}{\transparent{1}#2}}}

\newcommand\gbased{XPROA}
\newcommand\pbased{XPROB}

\newcommand\Tstrut{\rule{0pt}{2.1ex}}       % "top" strut
\definecolor{Gray}{gray}{0.9}
\begin{document}

\title{Explaining text classifiers through progressive neighborhood approximation with realistic samples %higher-diversity 
%Explaining with exemplars generated by boundary approaching*\\
%{\footnotesize \textsuperscript{*}Note%: Sub-titles are not captured in Xplore and
}

\author{
\IEEEauthorblockN{Yi Cai}
\IEEEauthorblockA{\textit{Dept. of Math. and Comp. Science} \\
\textit{Freie Universität Berlin}\\
Berlin, Germany \\
yi.cai@fu-berlin.de}
\and
\IEEEauthorblockN{Arthur Zimek}
\IEEEauthorblockA{\textit{Dept. of Math. and Comp. Science} \\
\textit{University of Southern Denmark}\\
Odense, Denmark \\
zimek@imada.sdu.dk}
\linebreakand
% \IEEEauthorblockN{Eirini Ntoutsi\IEEEauthorrefmark{1}\thanks{\IEEEauthorrefmark{1}Eirini Ntoutsi was affiliated with Freie Universität Berlin where most of the work was carried out.}}
\IEEEauthorblockN{Eirini Ntoutsi}
\IEEEauthorblockA{\textit{Research Institute CODE} \\
\textit{Universität der Bundeswehr München}\\
Munich, Germany \\
eirini.ntoutsi@unibw.de}
\and
\IEEEauthorblockN{Gerhard Wunder}
\IEEEauthorblockA{\textit{Dept. of Math. and Comp. Science} \\
\textit{Freie Universität Berlin}\\
Berlin, Germany \\
gerhard.wunder@fu-berlin.de}
}

\maketitle

\begin{abstract}
The importance of neighborhood construction in local explanation methods has been already highlighted in the literature.
And several attempts have been made to improve neighborhood quality for high-dimensional data, for example, texts, by adopting generative models.
Although the generators produce more realistic samples, the intuitive sampling approaches in the existing solutions leave the latent space underexplored.
To overcome this problem, our work, focusing on local model-agnostic explanations for text classifiers, proposes a progressive approximation approach that refines the neighborhood of a to-be-explained decision with a careful two-stage interpolation using counterfactuals as landmarks.
% Another novelty is that 
We explicitly specify the two properties that should be satisfied by generative models, the reconstruction ability and the locality-preserving property, to guide the selection of generators for local explanation methods.
% Sharing a similar idea, we further propose a method with the probability-based edition as an alternative to the generator-based solution due to the concern about the additional opacity introduced by the generator.
% Furthermore, due to the concern about the opacity of the generative models, we propose a method that implements progressive neighborhood approximation with probability-based editions as an alternative to the generator-based solution.
Moreover, noticing the opacity of generative models during the study, we propose another method that implements progressive neighborhood approximation with probability-based editions as an alternative to the generator-based solution.
The explanation results from both methods consist of word-level and instance-level explanations benefiting from the realistic neighborhood.
Through exhaustive experiments, we qualitatively and quantitatively demonstrate the effectiveness of the two proposed methods.
% The importance of the neighborhood for training a local surrogate model to  approximate the local decision boundary of a black box classifier has been already highlighted in the literature. 
% Several attempts have been made to construct a better neighborhood for high-dimensional data, like texts, by using generative autoencoders. However, existing approaches mainly generate neighbors by selecting purely at random from the latent space and struggle under the curse of dimensionality to learn a good local decision boundary.
% To overcome this problem, we propose a progressive approximation of the neighborhood using counterfactual instances as initial landmarks and a careful two-stage sampling approach to refine and generate factual as well as counterfactual instances in the neighborhood of the explaining target. 
% Our work focuses on textual data and our explanations consist of both word-level explanations from the original instance (intrinsic) and the neighborhood (extrinsic) and factual- and counterfactual-instances discovered during the neighborhood generation process that further reveal the effect of altering certain parts in the input text.
% Our experiments on real-world datasets demonstrate that our method outperforms the competitors in terms of usefulness and stability (for the qualitative part) and completeness, compactness, and correctness (for the quantitative part).
\end{abstract}


% TODO: check references from arxiv, replace them if possible
    % TODO: experiment: 01, 05 Add examples of explanation to appendix
% TODO: 05_extract, add discussion on feature entanglement
% TODO: 05_extract, rephrase the selection of factuals

\section{Introduction}
Advances in machine learning, deep learning in particular, prompt their applications in diverse real-world scenarios over past decades~\cite{pouyanfar2018survey}.
However, the rising model complexity and the exploding number of parameters~\cite{bernstein2021freely} keep drawing concerns over the transparency~\cite{goodman2017european} of the decision-making process for the systems steered by artificial intelligence.
Explainable AI (XAI) hence becomes a popular aspect that strives to uncover the underlying behaviors of black boxes wrapped by AI.

% There have been enormous efforts put into explaining models operating tabular data\cite{plumb2018model, ribeiro2016should, lundberg2017unified} and image data~\cite{petsiuk2018rise, selvaraju2017grad, chattopadhay2018grad}.
Despite the endeavor to explain models operating tabular data\cite{plumb2018model, ribeiro2016should, lundberg2017unified} and image data~\cite{petsiuk2018rise, selvaraju2017grad, chattopadhay2018grad}, most of them are not directly applicable to text classifiers owing to the unstructured nature of texts.
The relatively little attention leaves the field underdeveloped.
Thus, the main focus of this paper is to develop local model-agnostic explanation methods for text classifiers.
Approaches falling into this category first seek an interpretable local approximation of the explaining target with a simpler and, in most cases, linear surrogate model, then extract knowledge from the transparent local predictor as explanations for specific decisions instead of the whole black box because of its limited capacity.
A synthetic set (known as the neighborhood) consisting of samples close to the inquiry will be constructed during the explanation procedure for training the surrogate model in question.
% such a surrogate model.
Labeled by the black box, the synthetic data should reflect its behaviors.
In practice, the quality of the neighborhood dominates the upper bound of explanation quality.

To generate neighboring instances, perturbation of inputs~\cite{vstrumbelj2009explaining, strumbelj2010efficient} is a common choice.
It modifies the values of numeric features within a certain range or switches categorical features to some other values.
Again, the same task becomes much more challenging while dealing with textual data due to the lack of formal definitions of neighboring texts.
Some existing works~\cite{ribeiro2018anchors, ribeiro2016should} generate neighbors by randomly dropping words from the given text.
But perturbations ignoring inherent connections among the features (words) result in incomplete instances, which drag the focus of explanation out of the data manifold.
XSPELLS~\cite{lampridis2020explaining} proposes to generate neighboring texts in a latent space learned through a generative autoencoder.
Empowered by the generative model, it generates more realistic (semantically meaningful and grammatically correct) instances in comparison to the former approaches.
However, both groups of methods apply random sampling regardless of the construction space (either the original or latent), which leads to a spongy neighborhood.

% Our previous work~\cite{cai2021xproax} proposed \gbased{}~(local e\underline{X}plainer with \underline{PRO}gressive neighborhood \underline{A}pproximation), an explanation method for text classifiers that enhances explanation qualities through progressive neighborhood approximation with generative models.
% However, the opacity of the deployed generators powered by neural networks casts a shadow over the explanation process.
% As an extension of the previous conference paper~\cite{cai2021xproax}, this work addresses the concern by proposing \pbased{}~(local e\underline{X}plainer with \underline{PRO}bability-\underline{B}ased edition)\footnote{The source code is available at \url{https://github.com/caiy0220/XPROA-B}.} -- an alternative to the generator-based neighborhood approximation.
In this work, we propose two explanation methods\footnote{The source code is available at \url{https://github.com/caiy0220/XPROA-B}.} for text classifiers.
We first introduce \gbased{}~(local e\underline{X}plainer with \underline{PRO}gressive neighborhood \underline{A}pproximation), a local explanation method for text classifiers that implements neighborhood construction through a two-staged progressive interpolation relying on a generative model.
\gbased{}~enhances explanation qualities by refining the neighborhood, but the additional opacity of the deployed generator powered by neural networks casts a shadow over the explanation process.
Originating from this concern, we further propose \pbased{}~(local e\underline{X}plainer with \underline{PRO}bability-\underline{B}ased edition), which approximates the neighborhood through iterative editions on prototypes in a transparent manner.
As an extension of the conference paper~\cite{cai2021xproax}, the main contributions of the work are summarized below:
% This work extends our conference paper~\cite{cai2021xproax} in following aspects:
\begin{itemize}
    \item We explicitly specify two essential properties of generative models deployed for neighborhood construction in local explanations, namely the reconstruction ability and the locality-preserving property.
    \item We propose \pbased{}, an explanation method that constructs the neighborhood transparently through iterative editions on prototypes guided by local \textit{n}-gram context as an alternative to deploying generative models.
    \item We design an experiment that quantitatively evaluates the stability of explanation methods.
    \item The experimental results of dependency analysis expose the strong reliance of \gbased{}~on the generators, which underlines the concern about the opaque explanation process of generator-based explanation solutions.
\end{itemize}
% Besides, this version refines the experiment design of the previous work for quantitatively evaluating explanation methods following the three \textit{C}s of interpretability, namely correctness, completeness, and compactness.
Besides, this version refines the experiment design of quantitative explanation evaluation in the previous work for better alignment with the three \textit{C}s of interpretability, namely correctness, completeness, and compactness.
% In this work, we proposed two explanation methods\footnote{The source code is available at \url{https://github.com/caiy0220/XPROA-B}.} for text classifiers.
% Both enhance the explanation quality with progressive neighborhood approximation, which refines the neighborhood construction process.
% The main contributions of the work are summarized below:
% \begin{itemize}
%     \item We proposed \gbased{}~(local e\underline{X}plainer with \underline{PRO}gressive neighborhood \underline{A}pproximation), an explanation method that implements neighborhood construction through a two-staged progressive interpolation relying on a generative model.
%     \item Originating from the concern about the opacity of the generator powered by neural networks, we further proposed \pbased{}~(local e\underline{X}plainer with \underline{PRO}bability-\underline{B}ased edition). 
%     It constructs the neighborhood through iterative editions on prototypes guided by local \textit{n}-gram context as an alternative to deploying generative models.
%     \item Outputs of both methods are composed of word-level and factual-level explanations for comprehensive understanding.
%     \item We conducted detailed experiments for quantitative evaluation of explanation methods following the 3 \textit{C}s of interpretability, namely correctness, completeness, and compactness.
% \end{itemize}

The rest of the paper is organized as follows. 
In Section~\ref{sec:relatedWork}, we discuss state-of-the-art works regarding explainability.
Sections~\ref{sec:xproaxGen} and~\ref{sec:xproaxStats} introduce the proposed \gbased{}~and \pbased{}, respectively.
Afterward, Section~\ref{sec:explanationsMain} details the extraction of explanations from the surrogate model trained on the constructed neighborhood.
Exhaustive experiments are conducted in Section~\ref{sec:evaluation} for qualitatively and quantitatively evaluating the proposed methods.
Finally, we conclude the paper in Section~\ref{sec:conclusion}.

\section{Related work}
\label{sec:relatedWork}
The explainability of machine learning, especially deep learning, is drawing growing attention~\cite{danilevsky2020survey, linardatos2020explainable}. 
The existing solutions in this domain can be categorized as either \textit{global} or \textit{local} explanation methods depending on whether or not the explanation output is input-specific.
The former methods explain the target model as a whole at a global scope.
They provide model-level insights into the internal functions~\cite{puri2021cofrnets} and reveal the general correlations among features and the label~\cite{lundberg2020local}.
In contrast, the local ones analyze specific decisions of the target in detail, which leads to concrete and precise local explanations~\cite{ribeiro2016should, lundberg2017unified}.

As an alternative to the previous taxonomy, XAI approaches could be divided into the categories of \textit{model-specific} and \textit{model-agnostic} according to the access required on the to-be-explained model.
% Alternatively to the previous taxonomy, we could divide XAI approaches into the categories of \textit{model-specific} and \textit{model-agnostic} according to the access required on the to-be-explained model.
% As aforementioned, this work falls into the category of the local model-agnostic explanation method.
Model-specific methods generate explanations through in-depth analysis of the inner logic of the target~\cite{montavon2019layer}, for which full access to the model is a prerequisite.
Benefiting from the access, they could efficiently deliver faithful explanations. 
But the analysis procedure is highly specified since models handle data in diverse ways.
% Using neural networks as an example, which is the most common focus of model-specific explanations due to their popularity and complexity, existing solutions investigate the information flow through the neurons for understanding the decision-making process~\cite{shrikumar2017learning, montavon2017explaining}.
For example, existing model-agnostic approaches for neural networks investigate the information flow through the neurons for understanding the decision-making process~\cite{shrikumar2017learning, montavon2017explaining}.
Although neural networks in general share a basic idea, a single explainer from this category is usually eligible for a smaller subset of them.
LRP~\cite{bach2015pixel} is designed for analyzing the information flow on the dense and convolutional layers.
And adjustments to the backpropagation rules are mandatory before being applied to RNN~\cite{arras2017explaining} and attention mechanisms~\cite{ali2022xai}.

By comparison, model-agnostic methods only require permission to access the input and output of a model.
In principle, approaches of this kind are applicable to arbitrary black boxes~\cite{linardatos2020explainable}.
The loosened requirement makes them more flexible in security-sensitive scenarios, where models are opaque to outsiders (e.g., explanation methods).
Another difference between model-specific and model-agnostic methods is that the latter reveals the decision-making process indirectly through analysis of the causal relationship between the input features and the final outputs.
Neighborhood construction of the inquired instance comes in first of the analysis process.
A neighborhood consists of the synthetic data spatially close to the instance, on which the decision has been made.
The target model assigns labels to the generated neighboring instances to form a fully labeled dataset that reflects its local behavior.
The weights of the training sample could either be uniform~\cite{laugel2018defining} or in a distance-based manner~\cite{garreau2020explaining}.
Given the synthetic dataset, a surrogate model, chosen to be inherently transparent, will then be trained to mimic the local behaviors of the black box.
This local predictor will then serve as a proxy for deriving explanations.

As the first step, the neighborhood construction process plays a key role in model-agnostic explanation methods as it massively influences explanation quality.
The most intuitive solution is to perform a simple random perturbation~\cite{strumbelj2010efficient}, which alters feature values randomly within a predefined range. 
It iterates the slight fluctuation in the vicinity of the given input until the method collects sufficiently many neighbors.
Although this simple approach generates satisfactory neighbors for most data representations, it requires fine-tuning for more complicated forms of data, such as textual data in our case.
One difficulty of applying perturbations to text is how to modify the value of each feature (word). 
LIME~\cite{ribeiro2016should} and SHAP~\cite{lundberg2017unified} suggest \textit{word dropping} -- randomly removing/masking words from the input text. 
The solution is truly intuitive and has been shown to be effective. 
However, there still remain several problems:
\begin{itemize}
    \item The assumption of feature independence~\cite{vstrumbelj2014explaining} does not hold for texts because of syntax and semantics, random dropping results in incomplete and meaningless entries.
    \item The odd neighbors could shift the concentration away from the data manifold~\cite{chen2020true, ghalebikesabi2021locality}, where the explanation results might be unrepresentative of the model behaviors~\cite{frye2020shapley}.
    \item Perturbations are strictly limited by the length of the text, which becomes crucial when the input contains only a handful of words since the surrogate model will not have enough samples to approximate the black box.
\end{itemize}

Alternatively, XSPELLS~\cite{lampridis2020explaining} proposes to generate neighbors of texts with a variational autoencoder~\cite{bowman2016generating}. 
Preceding XSPELLS, ALIME~\cite{shankaranarayana2019alime} and ABELE~\cite{guidotti2019black} already deploy autoencoders for explaining models dealing with tabular data and images, respectively.
% Except for simple perturbations like word deletion, ABELE \cite{guidotti2019black} deploys a generative model for generating more "meaningful" neighbors with the consideration of the internal relationships between features. 
ALIME still generates neighbors in the original feature space, and the autoencoder only serves as part of the weighting function for measuring instance similarity.
But the neighborhood construction process of ABELE fully relies on the deployed adversarial autoencoder~\cite{makhzani2015adversarial}, a generator of synthetic images. 
XSPELLS fine-tunes the framework of ABELE to tackle the task of neighborhood generation for texts.
It constructs the neighborhood of an input text through random sampling in the latent space.
This way, it addresses the limitation of \textit{word dropping} by producing more realistic neighboring instances with a generative model.
However, sampling randomly cannot efficiently and effectively explore a high-dimensional latent space.
Moreover, its concrete choice of the generative model misses the two desired features of the generator for neighborhood construction under locality constraint~\cite{robnik2018perturbation, ghalebikesabi2021locality}, namely the reconstruction ability and the locality-preserving mapping.
% which leaves the generation strategy to be discussed

We address the limitations in the previous works with the proposed \gbased{}.
Although our method also adopts a generative autoencoder and achieves outstanding performance, we argue that the high dependency of explanation modules on opaque components, in this case, the generators, deserves further discussion.
Another solution -- \pbased{} -- from us shows that competitive performance can be achieved without involving extra obscurity.

% ! Besides, both ABELE and XSPELLS train the local predictor in the latent space.
% ! We argue that building the local predictor in a latent space can be a choice only if there are sufficient investigations on corresponding latent features.
% ! The proposed method extends XSPELLS by addressing the issues mentioned above. It differs from other approaches in the literature because of the progressive approximation of the neighborhood and the diversity-based \mbox{(counter-)}factual selection.

% the two properties for generator selection --> last paragraph of the first subsection

\section{Progressive neighborhood approximation with a generative autoencoder}
\label{sec:xproaxGen}
We introduce in this section the proposed explanation method \gbased{}.
Given $b(\cdot)$ as the function of a model to be explained, which takes a query text $x$ as the input and returns a class label\footnote{For simplicity, we assume a binary classification problem: $y\in \{+, -\}$.} $\hat{y}=b(x)$ as the classification result, the goal of the explainer is to find an explanation $\xi$ for the decision made by the target, i.e., $b(x)$.
Here we consider the model a black box, whose learnable parameters describing the function $b(\cdot)$ are wrapped and thus not accessible, but new queries on the model are unlimited.

As a model-agnostic method, \gbased{}~extracts explanations of the decision-making process indirectly through explainable approximations of the black box, which refers to a transparent surrogate model $m(\cdot)$ trained on a synthetic dataset $N$ that reflects the behavior of the target.
But the self-explainable constraint limits the capacity of such a surrogate.
% Due to the self-explainable constraint, a surrogate possesses limited capacities.
In order for the weak learner $m(\cdot)$ to be on par with the complex model $b(\cdot)$, the synthetic instances $N$ should carefully address the local neighborhood of an inquiry $x$, where a faithful approximation is achievable.
The requirement of generating spatially close samples as a compromise with the limited surrogate competence is called the \emph{locality constraint}.

\subsection{Locality-preserving neighborhood generation}
\label{sec:generativeModel}
Constructing qualified neighborhoods $N$ is the main challenge for explaining text classifiers.
Previous attempts, such as word dropping~\cite{ribeiro2016should} and corruption~\cite{iosifidis2019sentiment}, deliver instances with semantic and syntactic errors falling out of the manifold.
For generating more realistic neighbors aligning the manifold, we utilize a generative autoencoder for the construction process.

A generative autoencoder $G$ consists of a deterministic encoder $E:\mathcal{X}\rightarrow\mathcal{Z}$ that maps from the text space to the latent space and a probabilistic decoder $D:\mathcal{Z}\rightarrow\mathcal{X}$ reconstructing word sequences from their latent representations.
To prepare the generator for explanations, the autoencoder $G$ is trained in an unsupervised manner upon data $X_G\in\mathcal{X}$, which are sampled from the same domain as those fed to the black box (but please note that $X_G$ does not necessarily have intersections with the training set $X\in\mathcal{X}$ for the classifier).
During neighborhood construction, the prepared generator $G$ first maps the target instance $x$ to the latent space with the encoder and obtains its latent representation $z=E(x)$. 
Once encoding finishes, a set of neighboring vectors $Z$ is generated through manipulations on the pivot point\footnote{The latent space representation of the input $x$.} $z$ in the latent space, which are analogous to the process for numerical data.
% A set of latent vectors $Z$ is then constructed by manipulating the pivot point $z$ in the latent space the same as for numerical data.
After the latent space manipulations, the neighborhood $N$ of the inquiry $x$ for training the surrogate model can be derived using the decoder $N=D(Z)$.
With the prepared generative model, we enhance the semantic and syntax correctness of the neighboring texts.

The essential assumption of generators leveraged in neighborhood construction is the satisfaction of the \emph{reconstruction ability} and \emph{locality-preserving property}.
The former allows the generator to concentrate on the targeted instances.
And the latter means that the generator should map similar texts to nearby latent vectors, which ensures that neighbors discovered in the latent space are close to the inquiry in text form.
However, not all models possess the locality-preserving property~\cite{zhao2018adversarially}.
The risks of violating the locality constraint make them unsuitable for this scenario.
The generator adopted by this work, DAAE~\cite{shen2020educating}, learns locality-preserving latent representations by augmenting an adversarial autoencoder with a denoising objective, that it reconstructs original texts from their perturbed version.
Its adequately organized latent space geometry enables the generation of high-quality and similar synthetic texts for training the surrogate.
Besides, DAAE accomplishes the reconstruction task while VAE sacrifices its reconstruction ability due to the KL regularization term~\cite{song2019latent, dai2020usual}.
% This property of the adopted generator 
Being capable of reconstructing given inputs allows our explanation method to correctly locate the neighborhood given by the input $x$ rather than some others.

\subsection{Neighborhood generation with progressive boundary approximation}
\label{sec:neighborhoodGeneration}
Another main factor of latent space neighborhood construction is the manipulation strategy.
The choices range from pure random perturbation, locality-based sampling, and genetic algorithms to the proposed progressive neighborhood approximation.
Pure random perturbation~\cite{ribeiro2016should, lundberg2017unified, lampridis2020explaining} is the most intuitive solution, but it cannot produce compact neighborhoods and thus fails to focus on local decisions.
Locality-based sampling~\cite{laugel2018defining} addresses the problem by constraining sampling within a fixed-size hyper-sphere as specified by the closest counterfactual to the query instance $x$.
But since instances within the sphere are still sampled uniformly, the constructed set may not highlight the decision boundary.

The usage of a genetic algorithm~\cite{guidotti2018local} contributes to approaching local decision boundaries; however, the convergence guided by the fitness function harms the variety of the synthetic set. 
Moreover, its time cost is exceptionally higher than other solutions due to the genetic algorithm, which could limit its practical applications.
To overcome the limitations of the sampling approaches listed above and better approximate the local data structure surrounding $x$, we propose a progressive approximation of the neighborhood using landmarks to delimit the neighborhood boundary and by carefully interpolating within this boundary.

\begin{algorithm}[tb]
\caption{Neighborhood construction}
\label{alg:construct}
\begin{algorithmic}[1]
\renewcommand{\algorithmicrequire}{\textbf{Input:}}
\renewcommand{\algorithmicensure}{\textbf{Output:}}
\REQUIRE $x$: query instance; $L$: landmark set % landmark
\ENSURE  $N$: neighborhood of $x$
\STATE $C=\varnothing$
\WHILE{not $\mathrm{terminate}()$}
    \STATE $C_{\mathit{new}},~L_{\mathit{new}}=\mathrm{Approximate}(L, x)$
    \STATE $C=C\cup C_{\mathit{new}}$
    \STATE $L=L_{\mathit{new}}$
    % \STATE $N.append(N_{\mathit{new}})$
\ENDWHILE
\STATE $C=\mathrm{RemoveDuplicates}(C)$
\STATE $N=\mathrm{Closest}(C, n, n)$ \\ \COMMENT{\textit{output the $n$ closest instances for both classes}}
\RETURN $N$ 
\end{algorithmic} 
\end{algorithm}
The proposed approach (Algorithm~\ref{alg:construct}) constructs the neighborhood for a query through an iterative process, which achieves better neighborhood approximations over iterations as closer counterfactuals to $x$ are discovered and exploited.
Different from the previous attempts, which determine the neighborhood with a fixed hyperparameter, we initialize neighborhoods with landmarks from real data.
% The definition of the neighborhood for an instance depends on the corresponding data distribution and the decision boundary of the black box.
% It could vary when the to-be-explained target changes.
% Therefore, a fixed parameter defining the search space is not the optimal solution.
% Instead, we use a set of real instances (landmarks) to initialize the approximation process.
The initial landmarks are counterfactual instances from a retained corpus $X_L$, which is down-sampled from the generator training set $X_L\subseteq X_G$.
They help the construction process adapt to non-uniform data distribution.
% The usage of real data helps the construction process adapt to non-uniform data distribution.
More specifically, we use the \textit{k}-closest counterfactuals (the instances predicted by $b(\cdot)$ as of the opposite label of $x$) in the corpus $X_L$ depending on their latent distance to the pivot to initialize the landmark set $L$.
Thereafter, the neighborhood approximation (line 3) is performed iteratively on $L$ with two-staged interpolations until the termination criterion is satisfied (for example, when the approximation finds no closer counterfactual in successive rounds or the repeating count reaches its limitation).
In each iteration, the interpolated instances are archived in the candidate pool $C$ (line 4), and the landmark set is updated (line 5) with the newly generated counterfactuals (a subset of $C_{\mathit{new}}$).
Once the approximation terminates, we finalize the neighborhood $N$ containing $2n$ samples by picking the $n$ closest factuals and $n$ closest counterfactuals from the candidate pool without duplication (lines 7, 8).
% Once the approximation terminates, the final neighborhood $N$ with size $|N|=2\times n$ is extracted from the candidate pool by picking the $n$ closest non-duplicated factual and counterfactual instances (lines 7, 8).

Interpolation defined by \eqref{interpolate} is a commonly used operation in autoencoder-based text generation \cite{bowman2016generating, song2019latent, zhao2018adversarially}:
\begin{equation}
I(z_p, z_q)=\{z_i \mid z_i=z_p+i\cdot \frac{(z_p-z_q)}{s+1}, 0\leq i\leq s\} \label{interpolate}
\end{equation}
where $s\geq0$ is the number of interpolations between two poles.
The operation on latent vectors implements successive and meaningful text manipulations based on two selected prototypes -- $z_p$ and $z_q$.
The neighborhood approximation described in Algorithm~\ref{alg:approximate} is dominated by the two-staged interpolation, which consists of:
i) interpolation between pairs from the landmark set $L$ to allow for the exploration of the neighborhood and the variety of counterfactuals;
ii) interpolation between counterfactuals and the pivot point $z$ to allow for a more localized approximation of the decision boundary.

The approximation process takes as input the current landmark set $L$ and returns the updated landmarks $L_{\mathit{new}}$ along with the newly generated neighboring instances $C_{\mathit{new}}$. 
The \emph{first stage} (line 4) interpolates between a pair of randomly selected counterfactuals from $L$ (line 3).
The set of interpolated points $Z'$ from the first stage is labeled with the black box $\hat{y}_i=b(D(z_i))$.
In the \emph{second stage}, we interpolate between two sentiment polarities, namely between the target point $z$ and every instance $z_i\in Z'$ that satisfies $b(D(z_i))\neq b(x)$ (lines 7, 8).
All synthetic instances derived in this stage are decoded and added to the candidate pool for the final selection (line 11).
The set $L_{\mathit{new}}$ absorbs the closest counterfactuals from the newly generated candidates as landmarks for the next iteration (line 12).
The process repeats for $k$ times (lines 2, 13) to better explore the local space.
% The process repeats for $k$ times (lines 2, 13) allowing a better exploration of the constrained local space.
Note that the hyperparameter $k$ is the one for the initialization of the landmark set, so it remains the same size over iterations.

\begin{algorithm}[tb]
\caption{Neighborhood approximation}
\label{alg:approximate}
\begin{algorithmic}[1]
\renewcommand{\algorithmicrequire}{\textbf{Input:}}
\renewcommand{\algorithmicensure}{\textbf{Output:}}
\REQUIRE $x$: query instance; $L$: landmark set % landmark
\ENSURE  $C_{\mathit{new}}$: new neighbors; $L_{\mathit{new}}$: updated landmarks
\STATE $L_{\mathit{new}}, C_{\mathit{new}}=\varnothing,\varnothing$
\REPEAT
    \STATE $z_p, z_q = \mathrm{RandomlyDraw}(L, 2)$ % \\ \COMMENT{\textit{draw randomly 2 vectors from the landmark set}}
    \STATE $Z'=I(z_p,z_q)$ \\ \COMMENT{\textit{1st interpolation: between landmarks}}
    \STATE $Z=\varnothing$
    \FOR[\textit{2nd interpolation: between poles}]{$z_i\in Z'$}
        \IF{$b(D(z_i)) \neq b(x)$}
            % \STATE $Z.append(I(z_i, E(x)))$
            \STATE $Z = Z \cup I(z_i, E(x))$
        \ENDIF
    \ENDFOR
    \STATE $C_{\mathit{new}} = C_{\mathit{new}} \cup D(Z)$
    % \STATE $z_c = \mathrm{ClosestCounterfactual}(Z, 1)$
    \STATE $L_{\mathit{new}}.\mathrm{add}(\mathrm{ClosestCounterfactual}(Z))$
    % \STATE $C_{\mathit{new}}.append(Z)$    
\UNTIL{$k$ times}
\RETURN $C_{\mathit{new}}, L_{\mathit{new}}$ 
\end{algorithmic} 
\end{algorithm}

Through the iterative process, an optimal set of neighboring instances is discovered. 
The constructed neighborhood $N$ achieves higher diversity with the first stage interpolation and highlights the decision boundary between the pivot point and counterfactuals through the second stage.

\section{Neighborhood approximation with probability-based edition}
\label{sec:xproaxStats}
Although the generative autoencoder contributes to more realistic neighboring texts, building explanations upon another opaque model (current autoencoders are implemented mainly by neural networks with non-linear functions, which are usually considered not self-explainable) brings additional obscurity to the explanation process and limits the trustworthiness of the results.
Moreover, as demonstrated by our experiment (Section~\ref{sec:dependency}), the quality of explanations depends heavily on the concrete choices of the generator.  
The sensitivity to the opaque generative models makes the generator-based explanation methods vulnerable.
% the sensitivity of explanation quality to the generator capacity makes the explanation method vulnerable and impractical as there are no quantitative measures for the construction and selection of the generator.
To this end, we implement an alternative of the neighborhood approximation procedure with fully transparent probability-based editions -- \pbased{}, which still generate neighbors creatively in contrast to word dropping.

\subsection{Prototypes selection}
The general idea of \pbased{}~is to edit several counterfactual prototypes in a transparent and controllable way that progressively approximates the target instance.
Variants created through the editions form the neighborhood $N$ for surrogate training.
Inspired by the usage of landmarks for initializing the latent neighborhood in Section~\ref{sec:neighborhoodGeneration}, we also adopt a corpus $X_L\subseteq \mathcal{X}$ for prototype selection.
The prototypes are the top-$k$ counterfactuals according to their closeness to the input text.
There are multiple options for the metric of text similarity.
But to get rid of the opaque latent representations, we measure the proximity of text pairs by the cosine distance of their \textit{tf-idf} vectors~\cite{manning2008introduction} with the vectorizer fitted on $X_L$.
Albeit the \textit{tf-idf} representation biases towards less frequent terms, it helps balance the distribution of over-represented words (e.g., stopwords) and those under-represented.

During the implementation, we use the identical $X_L$ for both proposed methods.
The reason for naming the samples from the same set differently, namely prototype and landmark, is to distinguish the editions conducted by the two methods, where \pbased{}~applies changes to words directly in the original textual form while \gbased{}~performs arithmetic operations in the latent space.
% To satisfy the locality-preserving constraint, 

\subsection{Probability-based edition}
Given a prototype $\hat{x}=(w_0, w_1, ..., w_{l})$ and a word $w^*$, the probability-based edition finds the best fit for $w^*$ under the context of the prototype.
In other words, it seeks an operation for integrating $w^*$ into $\hat{x}$, which maximizes the likelihood of the manipulated version:
\begin{equation} \label{context}
    % \argmax_{i\leq j}P(\hat{x}^i_0, w^*, \hat{x}^{l+1}_j)
    \argmax_{i,j}P(\hat{x}^i_0, w^*, \hat{x}^{l+1}_j)
\end{equation}
% where $\hat{x}^j_i$ denotes a contiguous subsequence $(w_i, ..., w_{j-1})$ of the text $\hat{x}$.
where $\hat{x}^j_i=(w_i, ..., w_{j-1})$ is a contiguous subsequence of the text $\hat{x}$ (denoted by $\hat{x}^j_i\prec \hat{x}$) with its subscript and superscript indicating the indices of the starting and ending words, respectively.
The manipulation operation defined by $i$ and $j$, which satisfy $0\leq i\leq j\leq l+1$, could be either an insertion at position $i$ when $i=j$ or a replacement of the subsequence $\hat{x}^{j}_{i}$ otherwise.

Instead of taking the whole sequence into account, we concentrate on the local context described by \textit{n}-gram~\cite{jurafsky2000speech}. % consider bidirectional n-gram probably in the future.
The local context consists of the preceding and succeeding words.
Since the context is given by the prototype, the goal becomes:
% \begin{equation} \label{contextGram}
%     \argmax_{i\leq j}P(\hat{x}^{i}_{i-n}, w^*, \hat{x}^{j+n}_{j})
% \end{equation}
\begin{equation}
    \argmax_{i, j}P_{\mathit{pre}}(w^*|\hat{x}^{i}_{i-n})P_{\mathit{suc}}(w^*|\hat{x}^{j+n}_j)
\end{equation}
which is solvable in quadratic time.
Here, $P_{\mathit{pre}}$ and $P_{\mathit{suc}}$ denote the conditional probability of $w^*$ given the preceding and succeeding tokens, respectively:
\begin{equation*}
    P_{\mathit{pre}}(w^*|\hat{x}^j_i)=\frac{P(\hat{x}^j_i, w^*)}{P(\hat{x}^j_i)},~~P_{\mathit{suc}}(w^*|\hat{x}^j_i)=\frac{P(w^*, \hat{x}^j_i)}{P(\hat{x}^j_i)}
\end{equation*}
We estimate the conditional probabilities on the same corpus where the prototypes originate:
\begin{align}
    % P_{\mathit{pre}}(w^*|\hat{x}^{i}_{i-n})=&~\frac{|\{x|x\in X_L, (\hat{x}^{i}_{i-n}, w^*)\in x\}|}{|\{x|x\in X_L, \hat{x}^{i}_{i-n}\in x\}|} \\
    P_{\mathit{pre}}(w^*|\hat{x}^{i}_{i-n})=&~\frac{|\{x|x\in X_L, (\hat{x}^{i}_{i-n}, w^*)\prec x^k_l\}|}{|\{x|x\in X_L, \hat{x}^{i}_{i-n}\prec x\}|} \\
    P_{\mathit{suc}}(w^*|\hat{x}^{j+n}_{j})=&~\frac{|\{x|x\in X_L, (w^*, \hat{x}^{j+n}_{j})\prec x\}|}{|\{x|x\in X_L, \hat{x}^{j+n}_{j}\prec x\}|}
\end{align}
% where $\hat{x}^j_i\prec x$ denotes that $hat{x}^j_i$ is a contiguous subsequence of $x$.
The number of tokens $n$ defining the range of local context balances the trade-off between the correctness and creativity of the edition.
Due to the limited corpus size, here we choose \textit{unigram} ($n=1$) to simplify the generation process.
In practice, we take a minimal value of $\epsilon=\frac{1}{|X_L|+1}$ for $P_{\mathit{pre}}$ and $P_{\mathit{suc}}$ instead of $0$ to avoid the vanishment of probabilities.

Furthermore, to allow insertion and replacement at the beginning/end of a text, $n$ padding tokens \textlangle\textit{PAD}\textrangle~are appended separately to the head and tail of the input. 
The padding length relies on the range of local context defined by \textit{n}-gram.
For \textit{unigram}, the padding adds one on each end.
However, we find that the edition becomes biased toward the padding token for its significantly frequent appearance (possessed by all entries) over the dataset and consequently prefers to replace the entire prototype with the target word, i.e., $i=0$, $j=l+1$.
This abrupt change conflicts with the intention of applying gradual changes.
Thus, we update the goal with a penalty based on the edited length of the text:
\begin{equation} \label{eq:contrainedEdit}
    % \argmax_{i, j}P_{\mathit{pre}}(w^*|\hat{x}^{i}_{i-n})P_{\mathit{suc}}(w^*|\hat{x}^{j+n}_j)/e^{j-i} \\
    \argmax_{i, j}\frac{1}{e^{j-i}}P_{\mathit{pre}}(w^*|\hat{x}^{i}_{i-n})P_{\mathit{suc}}(w^*|\hat{x}^{j+n}_j)
\end{equation}
The soft constraint on edition cost encourages the selection of operations that affect fewer prototype components at each step.

\subsection{Neighborhood approximation through iterative edition}
With the probability-based edition constrained by edition length, we can gradually approach the target sentence through iterative manipulation on prototypes with words $w^*\in x$ (Algorithm~\ref{alg:pbEdit}).
The prototype pool is initialized by the \textit{k}-closest counterfactuals from the corpus (line 1) and updated by the edited version repeatedly (lines 6, 9).
To balance the distributions of the ingredients in $x$ over the constructed neighborhood, we generate equally for each word $\lceil\frac{k}{|x|}\rceil$ variants through editions on randomly chosen prototypes at a step (lines 4, 5).
During the iterative process, editions on prototypes are determined greedily according to the given ingredients as defined by \eqref{eq:contrainedEdit}.
The neighborhood set $N$ absorbs the newly generated variants from each iteration (lines 6, 8) with the repeated instances discarded.
% The repeated variants or ones with seldom contexts, i.e. $P_{\mathit{pre}}(w^*|\hat{x}^{i}_{i-n})P_{\mathit{suc}}(w^*|\hat{x}^{j+n}_j)=\epsilon^2$, are considered invalid and discarded, whereas the valid instances are added to the final neighborhood $N$.
The process repeats until enough neighboring texts have been collected (lines 2, 10).
\begin{algorithm}[tb]
\caption{Iterative probability-based edition}
\label{alg:pbEdit}
\begin{algorithmic}[1]
\renewcommand{\algorithmicrequire}{\textbf{Input:}}
\renewcommand{\algorithmicensure}{\textbf{Output:}}
\REQUIRE $x$: query instance;
\ENSURE  $N$: neighborhood of $x$;
\STATE $L=$ initPrototypes$(x, X_L)$
\REPEAT
    \STATE $L_{\mathit{new}}=\varnothing$
    \FOR{$w^*$ in $x$}
        \STATE $N'=$ editOnPrototypes($w^*, L, \lceil\frac{k}{|x|}\rceil)$)
        \STATE $L_{\mathit{new}}=L_{\mathit{new}}\cup N'$ 
    \ENDFOR
    \STATE $N=N\cup L_{\mathit{new}}$
    \STATE $L=L_{\mathit{new}}$
    % \STATE $C_{\mathit{new}}.append(Z)$    
\UNTIL{enough instances in $N$}
\RETURN $N$ 
\end{algorithmic} 
\end{algorithm}

For better coverage of neighborhoods, we try to edit all prototypes with every token in $x$ and keep the additional variants (for each token, the variants generated after the objective $\lceil\frac{k}{|x|}\rceil$ has been met) as candidates in our implementation.
They will be sampled as complements in case of early convergence that editions of the current epoch reach the same outcome and not enough prototypes have been gathered for the next epoch.

\section{Extraction of local explanations}
\label{sec:explanationsMain}
Explanation extraction from the neighborhood $N$ is independent of the neighborhood construction details.
Once the neighborhood $N$ is ready, the black box $b(\cdot)$ labels all instances in the set.
This fully labeled synthetic dataset reflecting the black box local behavior helps extract information for explanations.
In particular, we propose two explanation components:
i) word-level importance (Section~\ref{sec:explanationsFeatures}) and ii) factual and counterfactual instances identified during the neighborhood approximation process (Section~\ref{sec:explanationsCounterfactuals}).
% the decision boundary

\subsection{Word-level explanation}
\label{sec:explanationsFeatures}
First, a surrogate model $m(\cdot)$ is trained upon the labeled neighborhood $N$.
There are multiple choices for the surrogate model as long as the self-explainable requirement is satisfied.
% ~\cite{TBD} find reference for self-explainable requirement
We adopt a linear regression model for easier quantification of feature attribution.
Unlike XSPELLS~\cite{lampridis2020explaining}, we prefer building the surrogate in the original text space over the latent since the latter organized by a generator is less accessible for human understanding.
% TODO: talk out the latent space feature entanglement here
Considering the limited capacity of a linear regressor, texts are presented by bag-of-words (for the local vocabulary of $N$) in order to simplify the complexity of the classification task.
Sequential information vanishes as the trade-off between integrity and feasibility. 

The surrogate is trained on $N$ with the weighted square loss function defined as follows:
\begin{equation}
\label{lossfunc}
% \mathcal{L}(N, x)=\sum_{x_i\in N}exp(\frac{-\mathit{d_C}(x_i, x)^2}{\sigma^2})\cdot(b(x_i)-m(x_i))^2
\mathcal{L}(N, x)=\sum_{x_i\in N}exp(\frac{-\mathit{d_C}(x_i, x)^2}{\sigma^2})\cdot(b(x_i)-m(x_i))^2
\end{equation}
The first term in \eqref{lossfunc} computes the weight of a synthetic text $x_i$ based on its cosine distance $d_C$ to the target and the kernel width $\sigma$, which is a hyper-parameter that controls the influence of the weighting function.
And the second term is the square loss of the surrogate model $m(\cdot)$ on simulating the local behavior of the black box $b(\cdot)$.

% After training, the importance values of the features/words can be derived from the regression model.
After training, the weights of features learned through regression are derived as feature attributions.
In addition to the input words, those introduced by the creative neighborhood construction are also rated.
Therefore, we further split the word-level explanation into two parts:
i) \emph{intrinsic word} refers to words originating from the input; ii) \emph{extrinsic word} indicates external words involved during the generation process.
% i) \emph{intrinsic words}: words originate from the input; and ii) \emph{extrinsic words}: external words only appear in the neighborhood.
%As soon as the surrogate model is trained, feature importance can be smoothly exported as the word-level explanation. 
Feature attribution quantifies the importance of the corresponding feature to the inference process.
For intrinsic words, intuitively, the value reveals feature contribution to the inquired decision.
And extrinsic features from the neighborhood are incorporated to facilitate further explanation of the decision~\cite{ma2017salient, wang2011image}.
Extrinsic words with high attributions indicate important observations for inference in the same or similar contexts. 
Analogous to word deletion for intrinsic words, edition with extrinsic words could also illustrate the potential impacts of the corresponding features on the classification results via the resulting manual factual and counterfactual instances. 
% Although random insertion/replacement of words affects the prediction $b(\cdot)$ already, we adopt the probability-based edition determining the optimal operation for a better quality of the manual (counter-)factuals.

% The example presented in Table~\ref{tbl:editExample} demonstrates a counterfactual delivered by the edition with an extrinsic word. 
% The edition builds the connection between the input ``\emph{would not recommend.}'' and the extrinsic word ``\textit{definitely}'', it reveals that the adverb describing the word \emph{recommend} plays an important role in switching the black box decision.

% \begin{table}[tbp]
% \caption{An example of edition with an extrinsic word}
% \begin{center}
% \begin{tabular}{ll} 
% \hline
% \textbf{Input}: would not recommend . & $b(\cdot)$: \labelNeg\\
% \hline
% \multicolumn{2}{l}{\textbf{Extrinsic word}: \hlBlue[0.17]{definitely}} \\
% \textbf{Edition}: would \underline{definitely} recommend . & $b(\cdot)$: \labelPos\\
% \hline
% \end{tabular}
% \label{tbl:editExample}
% \end{center}
% \end{table}

\subsection{Factuals and Counterfactuals as explanation}
\label{sec:explanationsCounterfactuals}
Benefiting from the realistic neighboring texts, the synthetic samples themselves could also contribute to the understanding process as (counter-)factuals.
Analysis of the intersections and differences between factual and counterfactual instances would highlight the irrelevant and crucial components for the predictions.
Contrary to greedily picking the closest instances to the target, which results in a homogeneous (counter-)factual set,  our selection process considers diversity on top of closeness.
A diverse (counter-)factual set can better underscore the most contributing ingredients with a limited number of samples, which tend to be shared by factuals and altered in counterfactuals.

% TODO: check the following description
We construct the factual and counterfactual sets separately.
To form the counterfactual set $\Xi$, we pick the neighboring text successively until sufficiently many have been gathered.
Each time, the remaining neighbors are sorted in descending order according to their scores calculated by~\eqref{eqQuality}, and the set $\Xi$ absorbs the best instance.
\begin{equation}
\label{eqQuality}
    \begin{split}
        s_i=\lambda(1-\mathit{d_C}(x_i, x))+(1-\lambda)&\sum^{x_p\neq x_q}_{\Xi\cup\{x_i\}}\frac{\mathit{d_C}(\Delta x_p, \Delta x_q)}{(|\Xi|^2+|\Xi|)/2} \\
        &(\mathrm{where}~\Delta x_p = x_p - x)
        % &s_i=\lambda\cdot(1-\mathit{d_C}(x_i, x))+(1-\lambda)\cdot div(x_i, \Xi) \\
        % &div(x_i, \Xi)=\sum_{x'\in \Xi}\frac{\mathit{d_C}(\Delta x_i, \Delta x')}{|\Xi|}\mathrm{,~where}~\Delta x' = x' - x
    \end{split}
\end{equation} 
The score of a candidate is a linear combination of its cosine similarity to the target $x$ and the normalized diversity~\cite{gong2019diversity} of the counterfactual set if $\Xi$ takes $x_i$.
The parameter $\lambda\in[0,1]$ steers the relative importance of diversity during the selection.
For target instances that locate distant from the origin, diversities of vectors from their neighborhoods can be subtle.
Therefore, we take the measure on their differences to the target $x$ denoted by $\Delta x'$.
The ranking for the candidates has to be updated whenever $\Xi$ changes (after the acceptance of the new member at the end of each step), which explains the reason for picking samples in a successive manner.
The same process is also applied for preparing factuals.

\section{Evaluation}
\label{sec:evaluation}
We first show examples of explanations from different methods for qualitative comparison (Section~\ref{sec:qualitative}).
The primary objective of the experiments is to quantitatively evaluate the effectiveness of explanations in terms of correctness, completeness, and compactness (Section~\ref{sec:quantitative}), 
followed by the verification of explanation stability through texts under similar contexts (Section~\ref{sec:stability}).
Afterward, the sensitivity analysis regarding the major hyperparameters used throughout the tests is shown in Section~\ref{sec:parameter}.
Lastly, due to the concerns of adopting a black box (the generative model) for explaining another model, we design an experiment for analyzing the dependency on external resources, which exposes the potential risk of leveraging generative models in explanations (Section~\ref{sec:dependency}).
Details about the experimental setting follow in the upcoming section.

\subsection{Experimental setup} \label{sec:exp_setup}
\emph{Datasets}: We evaluated the approaches on two real-world datasets: Yelp reviews~\cite{shen2017style} and Amazon reviews polarity~\cite{zhang2015character}.
The \textit{Yelp} dataset consists of restaurant reviews, with a maximal text length equal to 16.
Possible labels of the reviews are either positive or negative.
The \textit{Amazon review polarity} dataset contains customer reviews about different products with ratings ranging from 1 to 5 stars~\cite{mcauley2013hidden}.
A succeeding work~\cite{zhang2015character} discarded all neutral reviews (rated as 3 stars) and re-defined the reviews with $\leq 2$ stars ($\geq$ 4 stars) as negative (positive, respectively) for matching a binary classification task. 
Compromising with the constrained capacity of the generative model used in the proposed framework, we focused on short texts and only used review titles in our experiments.
We split both datasets into four disjoint subsets with sizes 200K/20K/2K/4K, corresponding to the training set for the generative model and the training/validation/test set for the black box. 
Besides, we downsampled 20K instances uniformly from the generator training set $X_G$ as the corpus $X_L$ for the initialization of the landmark (prototype) set.
Although the generator training set is notably (mostly 10 times) larger than the remaining for ensuring the quality of the generator, the setting is feasible in practice as the adopted generative autoencoder is trained in an unsupervised manner.

\emph{Text classifiers}: 
In principle, a model-agnostic explainer is applicable to arbitrary black box $b(\cdot)$.
And for the sake of the evaluation, we trained a Random Forest (RF) and a Deep Neural Network (DNN), which are often referred to as black boxes for their non-linearity.
Regarding RF, we utilized the implementation from the \textit{scikit-learn} library with the number of weak learners set to 400, and a \textit{tf-idf} vectorizer is fitted on $X_L$ for preprocessing of raw texts.
As for DNN, we implemented an eight-layer DNN with \textit{Pytorch}. % Probably mention the link and version?
An LSTM layer~\cite{hochreiter1997long} locates in the center of the hidden layers, which captures the sequential information of texts; the rest are fully connected layers with a ReLU activation function.
% To train the DNN with texts of various lengths, we transformed raw texts into dense embeddings in the pre-processing stage with a tokenizer and a padder, the padder ensures that all dense embeddings share the same length.

\begin{table*}[tbp]
\caption{Details about datasets, black box models and generative model}
\begin{center}
\begin{tabular}{|c|c|c|c|c|c|c|c|c|} 
\hline
% \multicolumn{3}{|c|}{\textbf{Table Column Head}}
Dataset & Avg. length & Generative model & \multicolumn{3}{|c|}{Black box} & \multicolumn{2}{|c|}{Accuracy$^{\mathrm{a}}$} & $\mathcal{L}_{rec}$ $^{\mathrm{a}}$\\
\cline{4-6}
\cline{7-8} 
 & of texts & Training Set & Training set & Valid set & Test set & RF & DNN & DAAE \\
\hline
Yelp & 8.83 & 200K & 20K & 2K & 4K & 0.9113 & 0.9548 & 2.79 \\
\hline
Amazon & 8.47 & 200K & 20K & 2K & 4K & 0.7655 & 0.7795 & 3.93 \\ 
\hline
\multicolumn{6}{l}{$^{\mathrm{a}}$The reported values are performance on the test set for black boxes.}
\end{tabular}
\label{tbl:dataset}
\end{center}
\end{table*}

\emph{Generative model}:
We adopted DAAE, a symmetric generative autoencoder, as the generator for neighborhood construction.
The hyperparameter setting for the generative model follows the suggestion of the original paper~\cite{shen2020educating}.
Both encoder $E$ and decoder $D$ are one-layer bidirectional LSTM with 1024 units. 
The encoder takes as input a 512-dimensional dense vector from the embedding layer and projects the LSTM outcome to the latent space (bottleneck) with a size of 128 via a fully connected layer.
The decoder reconstructs texts from latent vectors following a reversed procedure.   % symmetric
The same setting applies to both datasets.
More details about the datasets, classifiers, and generative models can be found in Table~\ref{tbl:dataset}.
The reconstruction loss $\mathcal{L}_{rec}$ in the table is the cross entropy between the inputs and their reconstructions.
% The reported reconstruction loss is ... computed on the cosine similarity to the input text ...

\emph{Competitors}: 
We compared the proposed explanation methods, \gbased{}~-- relying on a generative model and \pbased{}~-- neighborhood construction with probability-based edition, to the following approaches in the experiments:
% We compared the following competitors throughout the experiments including both proposed methods:
\begin{itemize}
    \item \textbf{LIME}~\cite{ribeiro2016should}: the most widely used local explanation method that applies a simple random perturbation to the input $x$ for constructing its neighborhood. 
    \item \textbf{XSPELLS}~\cite{lampridis2020explaining}: a method generating neighbors for texts based on randomly sampling latent vectors and derives explanations from a decision tree built in the latent space.
    \item \textbf{ABELE}~\cite{guidotti2019black}: a generator-based explanation method with the latent space neighborhood constructed by a genetic algorithm.
\end{itemize}
During the tests, we adopted the same generative models for all competitors (if applied), such that the experimental results give more credit to the explanation procedures rather than the quality of the generators.
Note that the last method originally is designed for image classifiers, and we adapted ABELE for textual data by integrating its latent sampling approach into the framework of \gbased{}.
The main difference between the adapted ABELE and \gbased{}~is the replacement of the progressive neighborhood approximation with a genetic algorithm.
With the remaining components kept identical, comparing the two methods highlights the impacts of the sampling and counterfactual selection strategies on the resultant explanations.

\emph{Hyperparameters for neighborhood construction}:
The two determining hyperparameters in \gbased{}~are the interpolation step $s$ and the number of landmarks $k$, which are set to 10 and 20, respectively, for all test cases.
We also empirically selected 80 as the prototype set size $k$ for \pbased{}.
For both methods, the maximal neighborhood size $p$ is 400.
Further analysis of the impacts of hyperparameters is reported in Section~\ref{sec:parameter}.

\subsection{Qualitative evaluation} \label{sec:qualitative}
For qualitative comparison, we list explanations from different approaches for the decisions on selected instances and assess explanation quality with human-grounded evaluation~\cite{carvalho2019machine, danilevsky2020survey}.

Table~\ref{tbl:quality} presents the explanations regarding different inputs.
% More sample explanations for other inputs locate in the Appendix~\ref{TODO}.
LIME assigns importance scores to features in given texts as explanations, whereas the others, which construct the neighborhood in a meaningful and creative way, give explanations consisting of four parts:
i) feature attribution of \textit{intrinsic words}, which indicates their contributions to the predictions, 
ii) the most important \textit{extrinsic words} to the local decision boundary,
iii) the selected factual and iv) counterfactual examples.
Most of the competitors determine feature attribution through the weights learned by the surrogate model.
We illustrate the feature attribution of the intrinsic words through a saliency map for all competitors besides XSPELLS because it outputs in favor of relative frequencies of words in (counter-)factuals determined through a latent decision tree as word-level explanations (numbers in brackets).
The saliency map highlights words with colored backgrounds, with the color denoting the sentiment a word contributes to and the saturation indicating its importance.
Here, we use the blue (red) background referring to the positive (negative, respectively) sentiment.
For the extrinsic words, we report the top-ranked by their attribution as the important ones.

\begin{table*}[tbp]
\centering
\caption{Example explanations by different methods}
% \begin{center}
\begin{tabular}{c | p{6.85cm} p{8.15cm} } 
    \hline
    \rowcolor{Gray}
    \textbf{Input 1} & \multicolumn{2}{p{15.6cm}}{i love this story \hfill Dataset: \textbf{Amazon}, \textbf{Random Forest} $b(\cdot)$: {\textcolor{blue}{0.87}}\Tstrut}
    \\ \hline
    \\[-0.5em] \textbf{LIME} & 
    \multicolumn{2}{l}{\textbf{Saliency}: i \hlBlue[0.35]{love} this story}
    \\[0.5em] \hline \\[-1em] 
    \multirow{2}{*}[-3.5em]{\textbf{XSPELLS}} &
    % \multirow{2}{*}[-1em]{\textbf{XSPELLS}} &
    \textbf{Factuals}:\Tstrut & \textbf{Counterfactuals}:\Tstrut \\
    & 1) i love this song to the collection of it s wonderful & 1) i would recommend this book , but it s not like hepburn\\
    & 2) i love it and music brings the story of beautiful story & 2) i love this book so this is not that i was looking\\
    & 3) i am looking to love the garden to a beautiful story & 3) just terrible music and\\
    & 4) i love something & 4) $\langle unk\rangle$ just to ride out of this story with a smile!\\
    & 5) i m $\langle unk\rangle^{\mathrm{b}}$ loved this one and he wrote it for the plant & 5) if i didn t need to watch this, it s a show \\
    & \textbf{Common words in factuals}: & \textbf{Common words in counterfactuals}:\\
    & love (0.220), story (0.122), beautiful (0.049) & story (0.152), would (0.060), book (0.060)
    \\[0.2em] \hline \\[-1em] 
    \multirow{2}{*}[-3em]{\textbf{ABELE}} & 
    \textbf{Saliency}: i \hlBlue[0.12]{love} this \hlBlue[0.11]{story} &
    \textbf{Extrinsic words}: \hlRed[0.22]{worst}~\hlBlue[0.16]{best}\\
    & \textbf{Factuals}: & \textbf{Counterfactuals}: \\
    & 1) the best cd s classic classics! the beautiful & 1) the worst movie ever this movie will woody $\langle unk\rangle$ and felt as\\
    & 2) the best cd s classic classics! the beautiful & 2) the worst movie ever this movie will woody $\langle unk\rangle$ and felt as\\
    & 3) the best cd s classic classics! the beautiful & 3) the worst movie ever this movie will woody $\langle unk\rangle$ and felt as\\
    & 4) the best cd s classic classics! the beautiful & 4) the worst movie ever this movie will woody $\langle unk\rangle$ and felt as\\
    & 5) the best cd s classic classics! the beautiful & 5) the worst movie ever this movie and pierce s lewis was most
    \\[0.2em] \hline
    \\[-1em] \multirow{2}{*}[-3em]{\textbf{\gbased{}}} & 
    \textbf{Saliency}: i \hlBlue[0.29]{love} this story &
    \textbf{Extrinsic words}: \hlRed[0.08]{hate}\\
    & \textbf{Factuals}: & \textbf{Counterfactuals}: \\
    & 1) i love this book & 1) i buy this story\\
    & 2) i love the story & 2) i hate the story\\
    & 3) i love this book like the book & 3) buy the whole story\\
    & 4) i love this story of music.     & 4) i tried this movie\\
    & 5) i love this guitar  & 5) i own this guitar
    \\[0.2em] \hline
    \\[-1em] \multirow{2}{*}[-3em]{\textbf{\pbased{}}} & 
    \textbf{Saliency}: i \hlBlue[0.28]{love} this story &
    \textbf{Extrinsic words}: \hlRed[0.20]{poor} \hlRed[0.17]{terrible}\\
    & \textbf{Factuals}: & \textbf{Counterfactuals}: \\
    & 1) love this a story & 1) this story \\
    & 2) i love this story! & 2) love very poor story \\
    & 3) love this story i & 3) love story too boring \\
    & 4) this story i love & 4) love this world s worst story and movie... \\
    & 5) i love a story & 5) this story i
    \\[0.2em] \hline
    \multicolumn{3}{l}{$^{\mathrm{b}}$The generic token for words out of vocabulary.} \\
    \multicolumn{3}{l}{}\\[-0.3em] \hline
    \rowcolor{Gray}
    \textbf{Input 2} & \multicolumn{2}{p{15.6cm}}{the desserts were very bland . \hfill Dataset: \textbf{Yelp}, \textbf{DNN} $b(\cdot)$: {\textcolor{red}{0.99}}\Tstrut} 
    \\ \hline
    \\[-0.5em] \textbf{LIME} & 
    \multicolumn{2}{l}{\textbf{Saliency}: the desserts were very \hlRed[0.03]{bland} .}
    \\[0.5em] \hline \\[-1em] 
    \multirow{2}{*}[-3.5em]{\textbf{XSPELLS}}
    & \textbf{Factuals}:\Tstrut & \textbf{Counterfactuals}:\Tstrut \\
    & 1) the carrot was a average. & 1) the baked potatoes was a very sweet and a small thing. \\
    & 2) the only real german food. & 2) great hotel at this. \\
    & 3) the only offer of the mid $\langle unk\rangle$ houses in the last area. & 3) the mashed potato was good. \\
    & 4) the desserts were very bland. & 4) they had the certain menu. \\
    & 5) the hashbrowns were bland. & 5) the good donuts are the food. \\ 
    
    & \textbf{Common words in factuals}: & \textbf{Common words in counterfactuals}:\\
    & within (0.067), num (0.067), days (0.067) & atrocious (0.083), dvd (0.083), used: (0.083)
    \\[0.2em] \hline
    \\[-1em] \multirow{2}{*}[-3em]{\textbf{ABELE}} & 
    \textbf{Saliency}: the desserts \hlRed[0.05]{were} very \hlRed[0.34]{bland}. &
    \textbf{Extrinsic words}: \hlBlue[0.36]{good}\\
    & \textbf{Factuals}: & \textbf{Counterfactuals}: \\
    & 1) the desserts were very bland. & 1) the desserts were very good. \\
    & 2) the desserts were very bland. & 2) the desserts were very good either. \\
    & 3) the desserts were very bland. & 3) the desserts were very good either. \\
    & 4) the desserts were very bland. & 4) the desserts were very good either. \\
    & 5) the desserts were very bland. & 5) the desserts were very good either.
    \\[0.2em] \hline
    \\[-1em] \multirow{2}{*}[-3em]{\textbf{\gbased{}}} & 
    \textbf{Saliency}: the desserts \hlRed[0.10]{were} very \hlRed[0.43]{bland}. &
    \textbf{Extrinsic words}: \hlBlue[0.21]{good} \hlRed[0.20]{sausage} \hlRed[0.18]{rooms} \\
    & \textbf{Factuals}: & \textbf{Counterfactuals}:\\
    & 1) the ingredients were very bland. & 1) the desserts were very good.\\
    & 2) the desserts were very very bland. & 2) the deserts were also very tasty.\\
    & 3) the sausage were very good. & 3) the beers were very good.\\
    & 4) the desserts were very clean. & 4) the desserts were pretty good.\\
    & 5) the sides were very bland. & 5) the rooms were very good.
    \\[0.2em] \hline
    \\[-1em] \multirow{2}{*}[-3em]{\textbf{\pbased{}}} & 
    \textbf{Saliency}: the desserts \hlRed[0.09]{were} very \hlRed[0.33]{bland}. &
    \textbf{Extrinsic words}: \hlBlue[0.43]{great} \hlBlue[0.42]{delicious} \hlBlue[0.38]{excellent} \\
    & \textbf{Factuals}: & \textbf{Counterfactuals}: \\
    & 1) the desserts were very bland. & 1) the great desserts, were great food, very bland. \\
    & 2) desserts were the very bland. & 2) the desserts are very delicious too bland. \\
    & 3) the desserts were bland. & 3) the desserts were very good. \\
    & 4) desserts were very bland & 4) great fresh food, friendly service, bland and were the desserts to go \\
    & 5) were the very bland: desserts & 5) the were very excellent desserts.
    % & \makecell[l]{\textbf{Factuals}: \\
    % 1) the desserts were very bland. \\
    % 2) desserts were the very bland. \\
    % 3) the desserts were bland. \\
    % 4) desserts were very bland \\
    % 5) were the very bland: desserts} 
    % & \makecell[l]{\textbf{Counterfactuals}: \\
    % 1) the great desserts , were great food , very bland. \\
    % 2) the desserts are very delicious too bland. \\
    % 3) the desserts were very good. \\
    % 4) great fresh food, friendly service, bland and were the desserts to go \\
    % 5) the were very excellent desserts.}
    \\[0.2em] \hline
\end{tabular}
\label{tbl:quality}
% \end{center}
\end{table*}

The \textbf{first instance} $x$ to be explained is ``\textit{i love this story}'' -- a simple input from the Amazon dataset correctly classified as positive by the RF model.
%The instance has been classified as positive by the DNN black box model. 
As told by the saliency maps, the competitors agree that the term ``\textit{love}'' is the most contributing feature to the decision.
\gbased{}~further gives an extrinsic word ``\textit{hate}'', an antonym of ``love'', from the novel neighboring instances, which would affect the prediction negatively if involved in the current context.
The extrinsic words from ABELE and \pbased{}~demonstrate that sentimental adjectives describing the noun would also have prominent influences on the decisions under the current context.
Meanwhile, XPSELLS, which counts word frequencies, admits that ``love'' is the most contributing feature.
But it also outputs less relevant words.
Especially for the word ``story'', its frequent appearances in factuals as well as in counterfactuals reflect the truth that it, as a neutral term, barely affects the decisions in this case.
% Besides, the instance-level explanations from XSEPLLS are also more general in contrast to \gbased{}~and \pbased{}, which have better concentrations on the local context through the neighborhood approximation.
Besides, the instance-level explanations from XSEPLLS are also more general in contrast to our methods, which have better concentrations on the local context through the neighborhood approximation.
Through the selected (counter-)factuals, we observe that \gbased{}, powered by a generative autoencoder, is better at generating realistic samples for neighborhood construction.
The other method, \pbased{}, could still involve grammatical mistakes in the produced neighboring texts because of the limited local context and the greedy choice during the iterative edition process.

The \textbf{second instance} is ``\textit{the dessert is very bland .}'' selected from the yelp dataset with a negative label assigned by the DNN model.
For this input, XSPELLS fails to deliver useful information with generic words and instances.
Although LIME identified the term ``bland'' as the most contributing feature, the notably low importance score poorly matches the confident decision.
Combining the weak overall attribution given the high confidence and the comparably higher attribution in the first example (with a less confident decision), we suspect that the attribution scores by LIME are inconsistent while explaining various predictions.
On the other hand, the last three approaches agree that ``bland'' dominate the decision, and their listed (counter-)factuals agree with the observation. 
However, similar to the results for the first example, (counter-)factuals listed by ABELE are mostly identical.
The repeated instances reflect the collapsed neighborhood, amplified by the sparse textual space, during the evolution led by the genetic algorithm.
Another interesting observation that shouldn't be omitted is the two extrinsic words by \gbased{}, ``sausage'' and ``rooms''.
Intuitively, the two neutral terms should not leak information about the sentiment classification and remain less contributory.
But the third sentence in the factual set (falsely classified as negative) selected by \gbased{}~demonstrates the negative contribution of ``sausage''.
The same holds for ``rooms''.
Although the fifth counterfactual containing ``rooms'' is correctly classified as positive, its confidence score (0.56) is considerably lower than the other two analogous instances, i.e., the first (0.91) and the third (0.93), that only differ from it by the one specific word.
A deeper investigation into the training set uncovers the cause of the misuses.
Both words are uncommon in the training set (appear less than 20 times in the training corpus with 20K samples) with highly imbalanced distributions between the two classes, which contain the same amount of instances.
The word ``rooms'' appears twice more frequently in the negative samples than in the positive, and the same ratio for ``sausage'' is greater than 4.
Given this observation, we believe that the over-representations of the two terms in the negative samples mislead the model to use them falsely as shortcuts for the negative sentiment.
The higher imbalance of the term ``sausage'' also explains its stronger impact on the decision.

With the qualitative evaluation of the two selected examples, we demonstrate that the two proposed methods correctly identify the most contributing features to the corresponding decisions and how the more detailed results could enrich the information of the explanations.
The output from \gbased{}~on the second example, in particular, highlights the great potential of extrinsic words and instance-level explanations for not just understanding but also debugging.
In addition, through the comparison to the other two generative-based explanation methods, we underline the effectiveness and necessity of proper latent neighborhood construction during the generation of local explanations.
% In both examples, the two proposed methods correctly identify the most contributing features to the corresponding decisions. Moreover, the detailed explanations enrich the information about the decisions.
\subsection{Quantitative evaluation} \label{sec:quantitative}
Lacking ground truth, quantitative evaluation of explanation methods remains challenging without consent on how it should be conducted.
In this section, we quantified the competitors' performance following the three Cs of interpretability~\cite{silva2018towards, nauta2022anecdotal}, namely Correctness, Completeness, and Compactness.
\begin{itemize}
    \item \emph{Correctness}, also known as fidelity~\cite{guidotti2018survey}, refers to the faithfulness of explanations with respect to the target model $b(\cdot)$;
    \item \emph{Completeness} indicates the coverage of explanations on relevant observations;
    \item \emph{Compactness} requires explaining results to be concise by excluding irrelevant features.
\end{itemize}

For \textit{correctness}, we report the fidelity and the $R^2$ score of surrogate models, which quantify the performance of the corresponding explanation methods at imitating the target.
The similarity of the surrogate to the target dominates the upper bound of the explanation correctness -- one cannot interpret a model through an unfaithful proxy.

Second, we indirectly evaluate the \textit{completeness} of explanations with \textit{explanation-guided manipulation}, that is, editing the input texts and recording the change of classification results~\cite{lertvittayakumjorn2019human, arras2016explaining}.
Confidence drops of the predicted class after manipulations reflect the completeness regarding the coverage of relevant features.
% masked out with the token <UNKNOWN>
Specifically, the manipulation masks out positively contributing features and repeats negatively contributing ones.
Apart from intrinsic words, for methods that create novel neighbors during the explaining process, extrinsic words are also considered for manipulation (through the probability-based edition introduced in Section~\ref{alg:pbEdit}).
To prevent the edition from emptying the whole input, a threshold $\eta$ (we empirically set it to 0.1) is given, and only words with an importance score exceeding the threshold are considered.

Lastly, we adopt AOPC (area over perturbation curve)~\cite{samek2016evaluating}, the cumulative sum of confidence drops after each manipulation operation, as the metric for \textit{compactness}:
\begin{equation}
    AOPC_{x}=\frac{1}{l}\sum_{i=1}^l(b(x)-b(x^{(i)}))
\end{equation}
where $x^{(i)}$ denotes a variant of $x$, on which the manipulations with the top-$i$ features have been applied.
And $l$ denotes the total number of words to be edited (with attribution $\geq\eta$) given by the explanation.
Different from completeness, the order of editions matters in compactness evaluation.
They are applied sequentially according to the descending order of the relevant features with respect to their contribution magnitudes. % the absolute values of their importance scores
AOPC will give higher scores to compact explanations that assign more weights to relevant features, and vice versa.
% Averaged confidence drop over manipulations
The average confidence \underline{d}rop \underline{p}er \underline{m}anipulation (abbreviated as DpM) is also an intuitive and convenient measure for compactness.
However, it becomes less precise during the sequential manipulation procedure since the confidence drop is bounded by 1, and the impacts of lower-ranked features could be underestimated.
This measure biases towards methods that applies fewer manipulations and distorts the comparison of compactness, especially for the analysis conducted on polarized classifiers, where drops could easily saturate as the classifiers tend to give high confidence scores for their predictions regardless of the output labels.
We, therefore, prefer AOPC over averaged confidence drop as the metric for compactness but report both in the experiments.

All results of the designed experiments are presented in Table~\ref{tbl:effectiveness}.
The reported values are the averaged performance over the test sets followed by the standard deviation.
The most prominent observation is the flawless fidelity of XSPELLS as a measure of correctness.
The reason behind the exceptional performance is a different choice of surrogate model.
XSPELLS builds a decision tree for approaching the black box, which could always accurately separate given samples once the tree becomes deep enough, whereas the others use a linear regressor.
The different surrogate also explains the absence of the $R^2$ score.
Back to the group with the linear model, despite the expanded feature space due to the novel neighboring instances, the two proposed methods surprisingly outperform LIME in terms of correctness, especially for the $R^2$ score.
This observation indicates that both progressive approximation methods constructing realistic neighborhoods succeed in highlighting local decision boundaries, which consequently simplifies the imitation task for the linear surrogate.
\begin{table*}[tbp]
\caption{Quantitative evaluation following the three Cs, $\eta=0.1$}
% \begin{center}
\centering
\begin{tabular}{|c|l||c|c|c|c|c|c|} 
\hline
\multirow{2}{*}{Dataset \& Model}
& \multirow{2}{*}{Explainer} 
& \multicolumn{2}{c|}{\textit{Correctness}} & \textit{Completeness} & \multicolumn{2}{c|}{\textit{Compactness}} 
& \multirow{2}{*}{\makecell[c]{Time cost \\ (s)}}\Tstrut \\
\cline{3-7}
& & $R^2$ score & Fidelity & Confidence drop & \makecell{DpM} & AOPC & \Tstrut\\
\hline
\multirow{5}{*}{\makecell[c]{Amazon \& RF}}
 & LIME & 0.827 ± 0.206 & \textbf{0.963 ± 0.084} & 0.204 ± 0.181 & 0.201 ± 0.164 & 0.235 ± 0.155 & \textbf{0.113}\Tstrut\\
 & XSPELLS & - & \underline{1.000 ± 0.000} & 0.109 ± 0.198 & 0.075 ± 0.182 & 0.107 ± 0.157 & 14.740\\
 & ABELE & 0.869 ± 0.065 & 0.935 ± 0.042 & 0.235 ± 0.272 & 0.320 ± 0.220 & 0.328 ± 0.220 & 156.609\\
 & XPROA & \textbf{0.928 ± 0.044} & 0.942 ± 0.037 & 0.503 ± 0.256 & 0.316 ± 0.218 & 0.410 ± 0.204 & 16.592 \\
 & XPROB & 0.873 ± 0.070 & 0.938 ± 0.035 & \textbf{0.509 ± 0.228} & \textbf{0.329 ± 0.199} & \textbf{0.411 ± 0.172} & 1.019 \\
\hline
\multicolumn{8}{l}{}\\[-4pt]
\hline
\multirow{5}{*}{\makecell[c]{Amazon \& DNN}}
 & LIME & 0.880 ± 0.127 & 0.896 ± 0.110 & 0.343 ± 0.293 & 0.225 ± 0.193 & 0.331 ± 0.223 & 
\textbf{0.018}\Tstrut\\
 & XSPELLS & - & \underline{1.000 ± 0.000} & 0.106 ± 0.218 & 0.073 ± 0.198 & 0.102 ± 0.174 & 14.031 \\
 & ABELE & 0.874 ± 0.059	& 0.936 ± 0.040 & 0.281 ± 0.293 & 0.253 ± 0.261 & 0.273 ± 0.249 & 62.416 \\
 & XPROA & 0.\textbf{912 ± 0.053} & \textbf{0.938 ± 0.035} & 0.567 ± 0.285 & 0.239 ± 0.241 & 0.437 ± 0.234 & 4.551 \\
 & XPROB & 0.871 ± 0.078 & 0.903 ± 0.051 & \textbf{0.610 ± 0.239} & \textbf{0.260 ± 0.230} & \textbf{0.456 ± 0.201} & 0.808 \\
\hline
\multicolumn{8}{l}{}\\[-4pt]
\hline
\multirow{5}{*}{\makecell[c]{Yelp \& RF}}
 & LIME & 0.819 ± 0.186 & 0.924 ± 0.133 & 0.370 ± 0.298 & 0.301 ± 0.267 & 0.375 ± 0.254 & \textbf{0.101}\Tstrut\\
 & XSPELLS & - & \underline{1.000 ± 0.000} & 0.250 ± 0.325 & 0.142 ± 0.284 & 0.202 ± 0.256 & 13.424 \\
 & ABELE & 0.883 ± 0.059 & 0.943 ± 0.039 & 0.458 ± 0.349 & 0.341 ± 0.334 & 0.395 ± 0.298 & 156.649\\ 
 & XPROA & \textbf{0.906 ± 0.061} & \textbf{0.959 ± 0.039} & \textbf{0.751 ± 0.204} & 0.377 ± 0.315 & \textbf{0.591 ± 0.206} & 14.887\\
 & XPROB & 0.871 ± 0.085	& 0.915 ± 0.075 & 0.690 ± 0.217 & \textbf{0.387 ± 0.313} & 0.541 ± 0.218 & 0.921\\
\hline
\multicolumn{8}{l}{}\\[-4pt]
\hline
\multirow{5}{*}{\makecell[c]{Yelp \& DNN}}
 & LIME & 0.757 ± 0.222 & 0.954 ± 0.101 & 0.506 ± 0.424 & \textbf{0.410 ± 0.367} & 0.594 ± 0.293 & \textbf{0.017}\Tstrut\\
 & XSPELLS & - & \underline{1.000 ± 0.000} & 0.288 ± 0.397 & 0.169 ± 0.355 & 0.225 ± 0.299 & 14.388\\
 & ABELE & \textbf{0.867 ± 0.063} & 0.941 ± 0.039 & 0.448 ± 0.434 & 0.263 ± 0.370 & 0.334 ± 0.346 & 62.370\\
 & XPROA & 0.864 ± 0.071	& \textbf{0.958 ± 0.041} & \textbf{0.827 ± 0.299} & 0.320 ± 0.372 & \textbf{0.623 ± 0.278} & 4.321\\
 & XPROB & 0.798 ± 0.127	& 0.927 ± 0.045 & 0.819 ± 0.290 & 0.356 ± 0.372 & 0.574 ± 0.273 & 0.787\\
\hline
\end{tabular}
\label{tbl:effectiveness}
% \end{center}
\end{table*}
Also, the proposed methods achieve outstanding performance in compactness and correctness.
Even though \pbased{}~dumps the opaque but powerful generative model, it remains competitive with \gbased{}.
% manages to accomplish competitive performance compared to \gbased{}.
Furthermore, it achieves the highest DpM in all test cases except for the DNN model for the yelp dataset.
The exception is caused by the saturation of confidence change, as previously discussed.
For this test case, editions on a single word could already cause a significant change (a confidence drop greater than $0.5$).
Any subsequent actions will only hold back the overall compactness regardless of the actual importance of corresponding features if measured by DpM.
Therefore, LIME, the method with the fewest actions, comes out on top.
Regarding the other two generator-based competitors, ABELE acquires similar figures to LIME, whereas XSPELLS has a relatively modest performance.
Our adapted version of ABELE for textual data differs from \gbased{}~in the neighborhood construction approach.
And the gap between their explanation qualities underlines the effectiveness of the progressive approximation in latent space.
The completeness and compactness of explanations from XSPELLS disagree with its exceptional performance on correctness.
% mention entanglement of features for building surrogate in the latent space 
The more complicated surrogate model and the entanglement of latent space features obstruct the derivation of accessible explanations, although it better approaches the explaining target.
Admittedly, \gbased{}~and \pbased{}~benefit from the neighboring extrinsic words during the manipulation compared to LIME. 
However, the performances of XSPELLS and ABELE, which also involve novel words, show that the concrete choices of extrinsic words with opposite sentimental meanings should take into consideration the local context and can have a prominent influence.
In general, the results of the quantitative evaluation match the conclusion drawn from the previous qualitative analysis.

In the last column, we report the time complexity\footnote{The time costs are reported on an Intel i7-11800H CPU with a single Nvidia RTX3070 Max-Q GPU.} of explaining for all competitors.
The time cost of the listed explanation methods consists of two parts, neighborhood construction and requests for the black box outcomes.
LIME is the most efficient method among all competitors.
Its intuitive perturbation on the target input takes trivial time during explanations, so the time cost relies mainly on the black box efficiency for labeling the neighboring instances, which causes inconsistent performances while explaining different classifiers.
The same observations also apply to \gbased{}~and ABELE.
These two methods exhaustively construct neighborhoods to approach the local decision boundary with iterative processes that constantly query the target model.
The copious inquiries result in the massive dependency of their time costs on black box efficiency.
In fact, the three generator-based explainers in the middle are all time-consuming.
Deployment of another network for generating neighboring instances delays the derivation of explanations.
\pbased{}~mitigates the time cost by unloading the generative model.
But the more complicated generation scheme demands a higher computational cost,  which differs from the performance of LIME by an order of magnitude.
Considering the shallower classifiers adopted in the experiment, the running time of \pbased{}~, which relies more on the construction process during the tests, will approach LIME while explaining complex models.
Although the real-time constraint does not strictly apply to the domain of XAI, we still want to point out that the complexity of explanation methods could limit their potential applications in practice, e.g., debugging~\cite{bhatt2020explainable} or debiasing~\cite{kennedy2020contextualizing, cai2022power} machine learning models).
% Also, even though no critical real time requirement applies while seeking for explainability, time cost of explanation methods would still limit their potential applications (e.g. debugging~\cite{bhatt2020explainable}, debiasing~\cite{kennedy2020contextualizing, cai2022power} machine learning models).
\subsection{Stability} \label{sec:stability}
A stable explanation method ought to deliver similar explanations for similar inputs~\cite{carvalho2019machine}.
Regarding textual data, similar inputs should be semantically and syntactically close.
% the definition of similarity is still fuzzy, and
However, the limited number of similar text sets in natural corpora impedes the evaluation of explanation stability.
Inspired by text data augmentation~\cite{rudinger2018gender, dixon2018measuring}, which serves the same purpose of creating similar texts, we manually construct a test set with pre-defined templates as shown in Table~\ref{tbl:templates}.
Here, a test case is formed by five instances from the templates with the placeholders \textlangle\textit{ADJ}\textrangle~and \textlangle\textit{NOUN}\textrangle~replaced by a chosen adjective-noun pair.
For each test case, the feature attributions of the inserted words should remain similar due to the similar context.
Thus, we measure explanation stability by computing the attribution deviations for the two inserted words among the five synthetic instances.
\begin{table}[tbp]
    \caption{Templates for creating similar texts}
    \centering
    % \begin{center}
    \begin{tabular}{cl}
        \hline \\[-1.7mm]
        \multirow{5}{*}{\makecell[c]{\textbf{Templates}}} & 
        1. \textlangle\textit{ADJ}\textrangle~\textlangle\textit{NOUN}\textrangle \\
        \\[-2.3mm]
        & 2. Very \textlangle\textit{ADJ}\textrangle~\textlangle\textit{NOUN}\textrangle \\
        \\[-2.3mm]
        & 3. The \textlangle\textit{NOUN}\textrangle~is \textlangle\textit{ADJ}\textrangle\\
        \\[-2.3mm]
        & 4. A very \textlangle\textit{ADJ}\textrangle~\textlangle\textit{NOUN}\textrangle \\
        \\[-2.3mm]
        & 5. This is a very \textlangle\textit{ADJ}\textrangle~\textlangle\textit{NOUN}\textrangle \\
        \\[-2.3mm]
        \hline
    \end{tabular}
    % \end{center}
    \label{tbl:templates}
\end{table}

We chose the \textit{yelp DNN} model as the black box for the stability test.
The choice is reasoned by the more accurate and polarized model predictions, through which we expect the classification behaviors regarding the listed variants to be more stable.
We list the concrete choices of adjectives and nouns specified for the yelp DNN model in Table~\ref{tbl:wordTable}.
% The selection of words is specified for the \textit{yelp DNN} model.
% easier dataset, better performing classifier, lower risk of overfitting and more stable classifier performance --> easier & accurate estimation of stability
For each sentimental pole, we selected ten adjectives that receive the highest confidence scores for the corresponding class.
The nouns possess confidence scores close to 0.5 for both classes while using a single word as input for the classification.
\begin{table}[tbp]
    \caption{Choices of adjective and noun}
    \centering
    \begin{tabular}{cl}
        \hline \\[-1.7mm]
        \makecell[c]{\textbf{Negative}\\\textbf{adjectives}} & \makecell[l]{horrible terrible wrong awful disappointed \\ poor bland worst bad cheap}\\
        % \makecell[c]{\textbf{Negative}\\\textbf{adjectives}} & horrible terrible wrong awful disappointed poor bland worst bad cheap\\
        \\[-1.7mm]
        \hline
        \\[-1.7mm]
        \makecell[c]{\textbf{Positive}\\\textbf{adjectives}} & \makecell[l]{delicious amazing excellent loved fantastic \\ wonderful perfect fresh great best} \\
        \\[-1.7mm]
        \hline
        \\[-1.7mm]
        \makecell[c]{\textbf{Nouns}} & \makecell[l]{bread soup pizza food meal salad drink \\ dessert fish steak}\\
        \\[-1.7mm]
        \hline
    \end{tabular}
    \label{tbl:wordTable}
\end{table}
Enumerating all combinations of adjectives and nouns results in 200 test cases.
% The averaged performance over all test cases is presented in Table~\ref{tbl:stability}.

The performance of all competitors regarding the stability evaluation is presented in Table~\ref{tbl:stability}.
We excluded XSPELLS from the comparison as it cannot constantly assign importance scores to the intrinsic features, which impedes the stability analysis.
For all values shown in the table, we first compute the summary statistics (mean, deviation, and set similarity) for each test case, which we call the case statistics.
We then average all case statistics and report the means.
Using case deviation as an example, the computation of the average case deviation can be formally described as follows:
\begin{equation*}
    \overline{\delta} = \frac{\sum_{ADJ.}\sum_{N.}\delta(\mathrm{Templates}\left(adj., n.\right))}{|\{ADJ.\}|\times|\{N.\}|}
\end{equation*}
where $\mathrm{Templates}\left(\cdot,\cdot\right)$ denotes the set of variants created with the templates given an adjective and a noun.

The first column of Table~\ref{tbl:stability} gives the mean of case deviations on the classification outcomes.
This trivial figure does not refer to similar outputs regarding all incomes.
Instead, it indicates that the black box generally produces consistent predictions for instances from the same test cases (containing the same adjective-noun pair).
Similar inputs with similar predictions build the premise for the stability evaluation.
The third and fifth columns list the average contribution magnitudes of inserted words, and columns 4 and 6 present the means of case deviations.
% Columns 3-6 list the averaged attributions and deviations of the inserted words interpreted by various methods.
We split the words into two groups according to word class and average the case statistics for words belonging to the same group.
Since the words have different sentimental polarities, we use the modulus of attribution to avoid the collision of different signs in case means. % absolute values
Although the focus of the section is on stability, we want to point out that the case mean of the nouns gives a glimpse into the correctness.
The low attributions assigned to nouns (column 5) by both \gbased{}~and \pbased{}~match the ground truth that the neutral terms leak little information for the sentiment classification task, which indicates the faithfulness of their explanation outcomes.
Now returning to stability, the average case deviations of feature attributions are expected to be low for both word groups given the stable predictions (observed in column 1), which reflects the steadiness of the assigned importance scores across every five variants.
% Given the stable predictions, the averaged case deviations of feature attributions, for both word groups, from a stable explainer are expected to be low, which reflects the steadiness of the assigned importance scores across every five variants.
Closely followed by \pbased{}, ABELE reaches the lowest deviation of the importance scores assigned to the adjectives benefiting from the compact neighborhoods.
The diversity of the neighborhoods constructed by \gbased{}~results in the highest figure.
As for the nouns, \pbased{}~clearly outperforms the others with a value of nearly only one-third of the second lowest achieved by ABELE.
However, even the best method provides less steady explanations, especially considering the barely changed prediction outcomes.
A possible reason for the observation is that the expanding feature space (the additional words in the longer templates) distorts the scaling of the surrogate model.
\begin{table}[tbp]
\caption{Stability on the test cases, reported on the DNN model for the yelp dataset}
\label{tbl:stability}
\centering
% \begin{center}
\begin{tabular}{|c|c||c|c|c|c|c|} 
% m{2cm}
\hline
\rule{0pt}{8pt} 
\multirow{2}{*}{\makecell[c]{$\overline{\delta}_{pred}$}} & \multirow{2}{*}{\makecell[c]{Explainer}} & \multicolumn{2}{c|}{$\xi_{adj.}$} & \multicolumn{2}{c|}{$\xi_{n.}$} & \multirow{2}{*}{$\overline{S_C}$}\\
\cline{3-6}
& & $\mu$ & $\overline{\delta}$ & $\mu$ & $\overline{\delta}$ &\rule{0pt}{2.5ex}\\
\hline
% \multirow{4}{*}{0.0049} & LIME & 0.4412 & 0.1093 & 0.0431 & 0.0619 & 0.8188\Tstrut\\
% &ABELE & 0.1385 & \textbf{0.0616} & 0.0643 & 0.0546 & 0.8628\Tstrut\\
% &XPROA & 0.4413 & 0.1483 & 0.0346 & 0.0552 & 0.7758\Tstrut\\
% &XPROB & 0.5506 & 0.0818 & 0.0197 & \textbf{0.0192} & \textbf{0.8879}\Tstrut\\
\multirow{4}{*}{0.005} & LIME & 0.441 & 0.109 & 0.043 & 0.062 & 0.819\Tstrut\\
&ABELE & 0.139 & \textbf{0.062} & 0.064 & 0.055 & 0.863\Tstrut\\
&XPROA & 0.441 & 0.148 & 0.035 & 0.055 & 0.776\Tstrut\\
&XPROB & 0.551 & 0.082 & 0.019 & \textbf{0.019} & \textbf{0.888}\Tstrut\\
\hline
\end{tabular}
% \end{center}
\end{table}

To mitigate the scaling effect and to have a unified measure quantifying the stability, we report the \textit{set similarity} in the last column as the ultimate metric for explanation stability.
The set similarity is the average cosine similarity between feature attribution vectors $\langle\xi_{adj.}, \xi_{n.}\rangle$ in a test case.
LIME performs modestly according to the set similarity, which corresponds to its varying attributions.
With its neighborhood construction strictly limited by the length of the input text, the shortage of neighboring instances causes unfaithful explanations for the shorter inputs.
And this finally results in the instability of feature attributions under similar contexts.
Aligning with the observations on the feature attribution deviations, \pbased{}~provides the most stable explanations throughout all test cases.
This observation is foreseeable as the controlled generation process offers dense and realistic neighborhoods while granting the surrogate model to concentrate on the intrinsic features, which follow balanced distributions.
ABELE also achieves competitive performance in the stability evaluation.
We attribute this to the usage of the genetic algorithm.
It ensures the neighborhoods converge to certain subspaces for similar inputs over the evolution, so the surrogates built on homologous neighborhoods deliver similar explanations.
In contrast, \gbased{}, another generator-based explanation method, suffers from inconsistencies in explanation results as a side effect of neighborhood diversity.
The two-staged interpolation with better coverage of the diverse neighboring texts largely expands the feature space by introducing novel instances and thus affects the consistency of explanations on intrinsic words.

\subsection{Sensitivity to hyperparameters} \label{sec:parameter}
To study the influence of the hyperparameters on explanation quality, we repeated the quantitative evaluation with different hyperparameter choices for \gbased{}~and \pbased{}.
In \gbased{}, the two hyperparameters dominating the latent space exploration are the interpolation step $s$ and the number of landmarks $k$.
The neighborhood size is another factor in the neighborhood construction process.
But it is excluded from the analysis of hyperparameter sensitivity since the number of generated instances is highly dependent on the other two parameters and does not reach the limitation in most cases.
We picked the interpolation steps from 6 to 14 with an interval of 2 with the landmark set having a size of 20.
For the test on the landmark number, the values vary from 10 to 30 while keeping the interpolation step at 10.
As for \pbased{}, we tested its sensitivity to the choices of the neighborhood size $p$ (the population limitation) and the number of prototypes $k$.
These are the only two parameters that affect the construction process.
Compared to \gbased{}~(demonstrated in Table~\ref{tbl:effectiveness}), the less time complexity of \pbased{}~allows us to determine the hyperparameter choices on a larger scale to highlight their effects.
For the population limitation, we chose five values that increase exponentially from 100 to 1600 with the prototype quantity set to 80.
Similarly, we also picked a geometric sequence for $k$ while the population sticks to 400.

\begin{figure*}
    \centering
    \begin{subfigure}[b]{1.\textwidth}
        \centering
        \includegraphics[width=\textwidth]{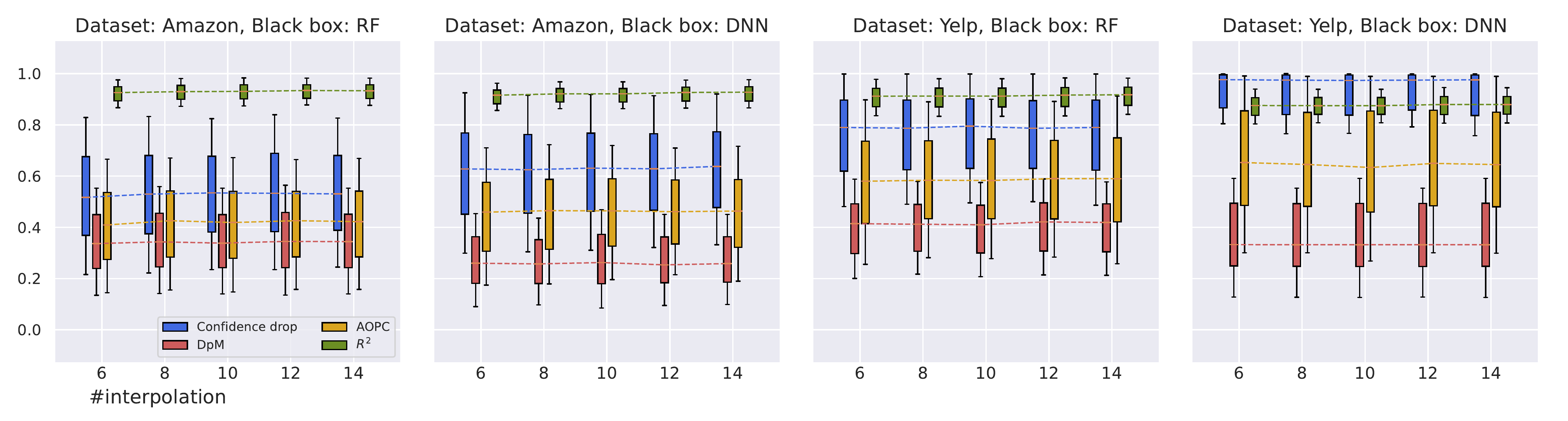}
        \vspace*{-8mm}
        \caption{Sensitivity to the interpolation interval, the number of landmarks is fixed to 20.}
        \label{fig:xproaIntpl}
    \end{subfigure}
    \vspace*{3mm}
    \begin{subfigure}[b]{1.\textwidth}
        \centering
        \includegraphics[width=\textwidth]{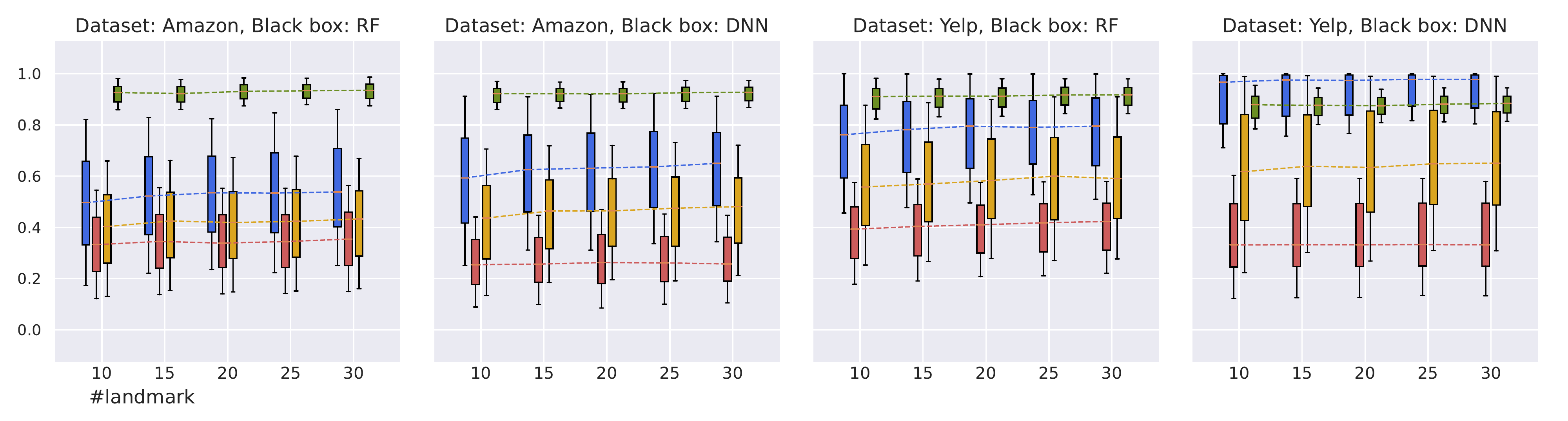}
        \vspace*{-8mm}
        \caption{Sensitivity to the number of landmarks, the population is fixed to 10.}
        \label{fig:xproaLm}
    \end{subfigure}
    \vspace*{-8mm}
    \caption{Sensitivity to hyperparameters of the neighborhood approximation in \gbased }
    \label{fig:xproa_sen}
\end{figure*}

For \gbased{}, Fig.~\ref{fig:xproaIntpl} shows that the interpolation intervals from the selected range have limited influence on the three Cs.
The reason for the observation is the continuous distribution in the latent space.
The increase in interpolation density only produces more duplicated instances, which are later removed during the neighborhood finalization.
On the other hand, the increasing size of the landmark set slightly improves the completeness and compactness, as visualized in Fig.~\ref{fig:xproaLm}.
The larger landmark set not only affects the initialization of the progressive approximation but also allows the usage of more landmarks during the iterative interpolation, which encourages latent space exploration and highlights the decision boundary.

\begin{figure*}
    \centering
    \begin{subfigure}[b]{1.\textwidth}
        \centering
        \includegraphics[width=\textwidth]{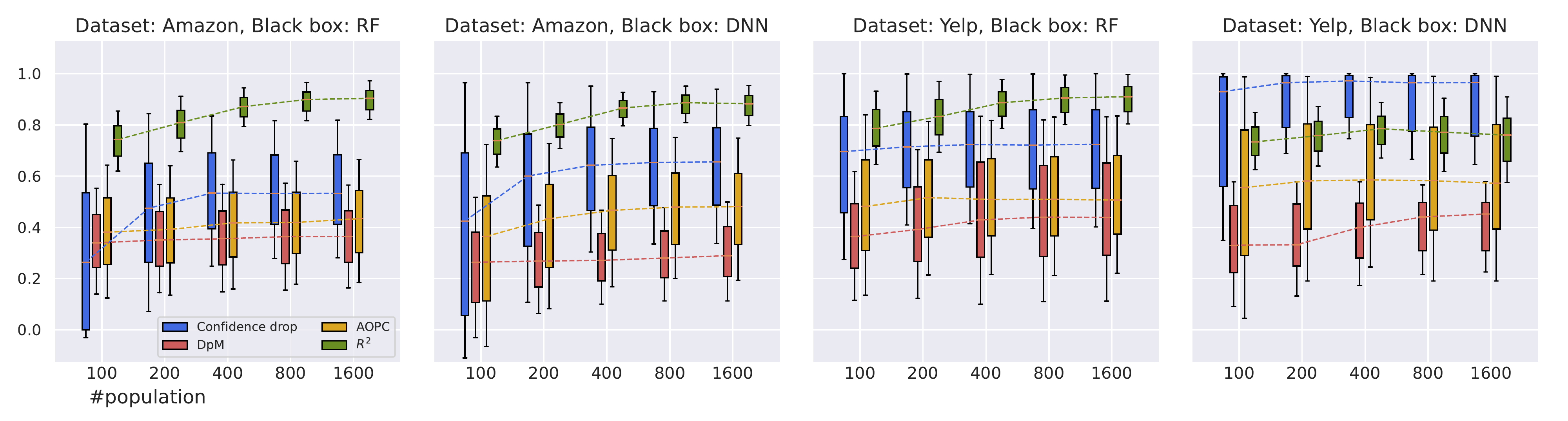}
        \vspace*{-8mm}
        \caption{Sensitivity to the neighborhood size, the number of prototypes is fixed to 80.}
        \label{fig:xprobPopu}
    \end{subfigure}
    \vspace*{3mm}
    \begin{subfigure}[b]{1.\textwidth}
        \centering
        \includegraphics[width=\textwidth]{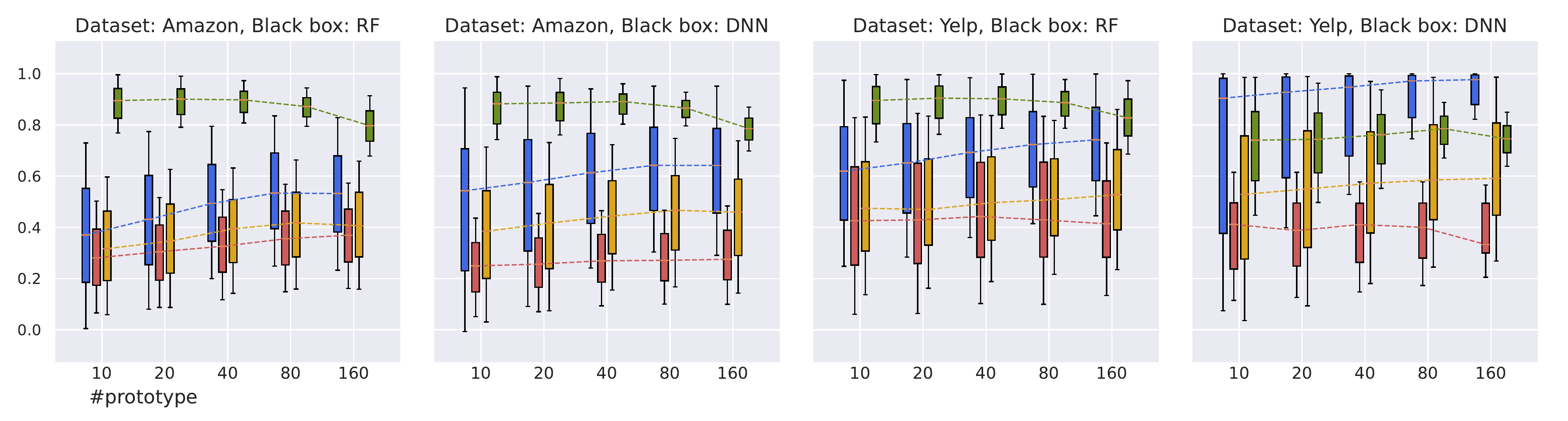}
        \vspace*{-8mm}
        \caption{Sensitivity to the number of prototypes, the population is fixed to 400.}
        \label{fig:xprobLm}
    \end{subfigure}
    \vspace*{-8mm}
    \caption{Sensitivity to hyperparameters of the neighborhood approximation in \pbased }
    \label{fig:xprob_sen}
\end{figure*}

Similar changes in confidence drop and in AOPC hold for \pbased{}~with the expansion of the prototype set (Fig.~\ref{fig:xprobLm}), which also affects the initialization along with the intermediate prototype selection.
Partly owing to the larger gaps between the hyperparameter choices, the increasing tendencies are more significant.
The changes in completeness suggest that more relevant features are filtered out and involved in manipulations.
And for the two classifiers trained on the yelp dataset, the saturation of confidence drop holds their DpMs back, which aligns with our argument in Section~\ref{sec:quantitative} about having AOPC as another compactness metric.
In contrast to the changing tendency of completeness, the dropping $R^2$ scores suggest that an excessive amount of prototypes can violate the locality constraint.
The expanding prototype set includes additional parts of the decision boundary and finally introduces non-linearity when it reaches far enough. 
But the explanation qualities remain steady even with a badly chosen $k$ ($=160$) because of the neighborhood approximation and the weighted loss function.
The former tightens the neighborhood, and the latter encourages the surrogate to concentrate on the closer samples by assigning them higher weights.
% ----------------------------
Meanwhile, Fig.~\ref{fig:xprobPopu} illustrates that enriching the neighborhood set also lays positive effects on explanation qualities.
According to all listed metrics, the performances improve rapidly with the growing population, especially when the number is low, and the upward trend slows down once $p$ reaches 400.
The observation indicates that explanations become precise when the population is large enough for proper coverage of neighborhoods.
After that, enrichment of neighboring samples would not further promote explanation qualities.
% Three testing cases out of the four obey the preceding description, but with one exception, whose completeness and AOPC experience a decline in the range between 400 and 1600.

% The same choice of hyperparameters is applied to explaining different classifiers for the quantitative evaluation.
% But the results of this section show that the optimal values are classifier specific.
Although we applied the same hyperparameter setting for the remaining parts of the experiments, the plots in Fig.~\ref{fig:xproa_sen} and Fig.~\ref{fig:xprob_sen} show that the optimal choices are not identical under different settings.
In reality, the best hyperparameter choice depends on the data distribution and the decision boundary of the explaining target.
% It would be time-consuming if the goal is determining optimal neighborhoods.

\subsection{Dependency on external resources} \label{sec:dependency}
% \subsection{Dependency on generative model} \label{sec:dependency}
As the last part of our experiments, we investigated in this subsection the dependencies of the two proposed methods on the external resources they utilized as knowledge for improving neighborhood quality.

Like many other generator-based explanation methods~\cite{frye2020shapley, lampridis2020explaining, guidotti2019black}, \gbased{}~determines the neighborhood of inputs in the latent space, which fully relies on the adopted generator.
Despite their importance in neighborhood constructions, there is yet little study on the impacts of generative models on explanation quality.
Originating from the concern about the opacity of the generators, we investigate the dependency of explanation quality on the generative models and discuss the potential risks of the generator-based solutions for XAI.
The capacity of a generative model is directly related to its ability to generate and reconstruct.
% The first term determines the quality of synthetic instances, and the second decides whether the generated neighborhood satisfies the locality constraint.
In order to uncover the correlation between generator capacity and explanation quality, we alter the size of the autoencoder bottleneck with the remaining components of \gbased{} unchanged and perform the quantitative evaluation on the \textit{amazon DNN} model.  % TODO: add remaining to Appendix
Furthermore, as discussed in Section~\ref{sec:generativeModel}, latent space geometry also plays a crucial role in the locality-preserving generation.
Therefore, we also replace the chosen DAAE with VAE~\cite{bowman2016generating}, a well-known generative autoencoder, and compare their performances with various bottleneck sizes.
DAAE differs from VAE mainly in the training scheme, it applies noises to the training samples for better latent space geometry and reconstruction ability.
% The major difference between DAAE and VAE is the training scheme.
For a valid comparison, the remaining network structures of DAAE and VAE are identical and remain constant during the test.

% ----------------- DAAE -----------------
The chosen bottleneck sizes are listed in the second column of Table~\ref{tbl:dependency}, followed by the corresponding reconstruction losses in column 3 reported on the test set.
For DAAE, the reconstruction loss correlates negatively with the size of the bottleneck.
The drop is heavier for the narrower bottlenecks and becomes trivial when its size approaches the capacity of the preceding and succeeding layers.
The lower reconstruction loss suggests a better fit for the locality constraint as the generator can better concentrate on the given instance.
% The properly organized latent space makes sure the synthetic texts sampled in the latent neighborhood are close to the target.
Not surprisingly, the figures for completeness (column 4) and compactness (columns 5 and 6) of explanations deploying DAAE grow accordingly when the generator capacity improves.
\begin{table}[tbp]
\caption{Dependency of \gbased{} on generative models, reported on the amazon DNN model}
\label{tbl:dependency}
\centering
% \begin{center}
\begin{tabular}{|c|c|c||x{0.9cm}|x{0.9cm}|x{0.8cm}|} 
\hline
Generator & Bottleneck & $\mathcal{L}_{rec}$ & C. D.$^{\mathrm{c}}$ & \makecell{DpM} & AOPC\Tstrut\\
\hline
\multirow{4}{*}{DAAE} 
& 32 & 11.77 & 0.464 & 0.229 & 0.373\Tstrut\\
\cline{2-6}
& 64 & 5.47 & 0.528 & 0.235 & 0.411\Tstrut\\
\cline{2-6}
& 96 & 4.18 & 0.557 & 0.236 & 0.435\Tstrut\\
\cline{2-6}
& 128 & 3.93 & \textbf{0.567} & \textbf{0.239} & \textbf{0.437}\Tstrut\\
\hline
\hline
\multirow{4}{*}{VAE} 
& 32 & 25.23 & 0.329 & 0.226 & 0.289\Tstrut\\
\cline{2-6}
& 64 & 25.72 & 0.268 & 0.196 & 0.235\Tstrut\\
\cline{2-6}
& 96 & 27.00 & 0.296 & 0.215 & 0.264\Tstrut\\
\cline{2-6}
& 128 & 27.75 & \textbf{0.332} & \textbf{0.231} & \textbf{0.299}\Tstrut\\
\hline
\multicolumn{6}{l}{$^{\mathrm{c}}$confidence drop}
\end{tabular}
% \end{center}
\end{table}
% ----------------- VAE --------------------
On the contrary, the increasing capability of VAE does not necessarily improve its performance on reconstruction due to the organization of the latent space~\cite{zhao2018adversarially}.
Even though the generator provides more realistic samples with the growing bottleneck size, the violation of the locality constraint results in the independence of explanation quality from generator capacity.
\begin{table}[tbp]
\centering
\caption{A failed case by \gbased{} with VAE}
% \begin{center}
\begin{tabular}{p{3.95cm}p{3.95cm}} 
    \hline \rowcolor{Gray}
    \multicolumn{2}{m{8.35cm}}{\textbf{Input}: stay the hell away from this book \hfill \textbf{Amazon}, \textbf{DNN} $b(\cdot)$: {\textcolor{red}{0.86}}\Tstrut} \\
    \hline \\[-0.5em]     
    % \textbf{LIME} & \hlRed[0.27]{stay} the \hlRed[0.07]{hell} \hlRed[0.07]{away} from this book\Tstrut
    \multicolumn{2}{m{8.35cm}}{\hlRed[0.27]{stay} the \hlRed[0.07]{hell} \hlRed[0.07]{away} from this book \hfill \textbf{LIME}\Tstrut}
    \\[0.5em] \hline \\[-0.5em] 
    \multicolumn{2}{m{8.35cm}}{\hlRed[0.32]{stay} the \hlRed[0.17]{hell} \hlRed[0.14]{away} from this book \hfill \textbf{\gbased{}~(DAAE-128)}\Tstrut}\\
    \scriptsize \textbf{Factuals}: & \scriptsize \textbf{Counterfactuals}:\Tstrut\\[-0.1em]
    \scriptsize \makecell[l]{ 
    1) stay the night it with this book!\\
    2) stay the hell away for this book\\
    3) stay the critics away in this book} &
    \scriptsize \makecell[l]{
    1) the far side away from this book\\
    2) enjoy the movie away from this book\\
    3) why the hell do schools this book}
    \\[0.5em] \hline \\[-1em] 
    \multicolumn{2}{m{8.35cm}}{stay the hell away from this \hlBlue[0.10]{book} \hfill \textbf{\gbased{}~(VAE-128)}\Tstrut}
    \\
    \scriptsize \textbf{Factuals}: & \scriptsize \textbf{Counterfactuals}:\Tstrut\\[-0.1em]
    \scriptsize \makecell[l]{
    1) where s roy with the rest?\Tstrut\\
    2) where is jamie lee s a?\\
    3) where s to the point?} &
    \scriptsize \makecell[l]{
    1) where s roy with my mind\\
    2) why is so many of these songs?\\
    3) where s the new wave}
    % \multicolumn{2}{l}{\makecell[l]{\textbf{Factuals}:\Tstrut\\
    % 1) where s roy with the rest? \hfill 2) where is jamie lee s a?\\
    % 3) where are the pictures to make this?\\
    % 4) where s $\langle unk\rangle$ at the end? \hfill
    % 5) where s to the point?\\
    % \textbf{Counterfactuals}:\Tstrut\\
    % 1) where s roy with my mind \hfill 2) why is so many of these songs?\\
    % 3) why did i have these types of my life?\\
    % 4) where s the new wave \hfill 5) where s $\langle unk\rangle$ when in the movie}} 
    \\ \hline 
\end{tabular}
\label{tbl:fail_example}
% \end{center}
\end{table}
Table~\ref{tbl:fail_example} presents a failure of \gbased{}~with a generator unsatisfying the assumptions on the reconstruction ability and locality-preserving property.
According to the (counter-)factuals in the last entry, failing to concentrate on the given context (because of the high reconstruction loss) leads to the ignorance of multiple ingredients in the input, which causes a notably worse performance.
Here, the explainer equipping VAE also reaches its best performance with the widest bottleneck.
But the narrowest bottleneck (with a size equal to 32) achieving similar results in all metrics suggests that this tends to be a coincidence.

\begin{table}[tbp]
\caption{Dependency of \pbased{} on the prototype corpus $X_L$, reported on the amazon DNN model}
\label{tbl:dependency_corpus}
\centering
\begin{tabular}{|c||x{0.9cm}|x{0.9cm}|x{0.8cm}|} 
\hline
Corpus size & C. D. & \makecell{DpM} & AOPC\Tstrut\\
\hline
2k & 0.539 & 0.257 & 0.441\Tstrut\\   % 0.573 & 0.266 & 0.467\Tstrut
\hline
5k & 0.593 & 0.266 & 0.472\Tstrut\\
\hline
10k & 0.595 & 0.261 & 0.472\Tstrut\\
\hline
20k & 0.607 & 0.266 & \textbf{0.473}\Tstrut\\
\hline
40k & \textbf{0.609} & \textbf{\textbf{0.267}} & 0.472\Tstrut\\
\hline
80k & 0.606 & 0.266 & 0.464\Tstrut\\
\hline
\end{tabular}
\end{table}
We also experimented with \pbased{}~for its dependency on the prototype corpus (Table~\ref{tbl:dependency_corpus}).
Analogous to the previous method, the explanation quality of \pbased{}~correlates positively with the external knowledge guiding the construction of neighborhoods.
The performance stabilizes when the corpus size becomes large enough to represent the distribution of the target domain.
For the selected classifier and dataset, it is $|X_L|\geq 5k$.
Compared to the generator-based solution, the one with the probability-based edition is less sensitive to the change of external resources.

% ====> A brief summary for the section <===
Through the sensitivity of explanations to particular choices of generators, we illustrate the heavy reliance of \gbased{}~on the generative model as well as the necessity of fulfilling the two properties mentioned in Section~\ref{sec:generativeModel}.
The significant correlation between reconstruction ability and explanation quality suggests that reconstruction loss should be considered one of the selection criteria for the generative model.
Meanwhile, the gap between the performances of explainers adopting DAAE and VAE emphasizes the importance of latent space geometry, which is unfortunately challenging to investigate due to its complexity and non-linearity.
% The dramatic changes in explanation quality corresponding to the modified training scheme further expose the vulnerability of the generator-based explanation methods.
% The reliance on the latent space that we know little about undermines the trustworthiness of the derived explanations.
And we argue that the reliance on such inaccessible latent space may undermine the trustworthiness of the derived explanations.

% \begin{table*}[tbp]
% \caption{Dependency of \gbased{} on generative models, reported on the DNN model for the amazon dataset}
% \label{tbl:dependency}
% \centering
% % \begin{center}
% \begin{tabular}{|c|c|c||c|c|c|} 
% \hline
% Generator & Bottleneck & $\mathcal{L}_{rec}$ & Confidence drop & \makecell{DpM} & AOPC\Tstrut\\
% \hline
% \multirow{4}{*}{DAAE} 
% & 32 & 11.77 & 0.464 ± 0.341 & 0.229 ± 0.258 & 0.373 ± 0.281\Tstrut\\
% \cline{2-6}
% & 64 & 5.47 & 0.528 ± 0.302 & 0.235 ± 0.244 & 0.411 ± 0.249\Tstrut\\
% \cline{2-6}
% & 96 & 4.18 & 0.557 ± 0.280 & 0.236 ± 0.241 & 0.435 ± 0.236\Tstrut\\
% \cline{2-6}
% & 128 & 3.93 & \textbf{0.567 ± 0.285} & \textbf{0.239 ± 0.241} & \textbf{0.437 ± 0.234}\Tstrut\\
% \hline
% \hline
% \multirow{4}{*}{VAE} 
% & 32 & 25.23 & 0.329 ± 0.364 & 0.226 ± 0.297 & 0.289 ± 0.303\Tstrut\\
% \cline{2-6}
% & 64 & 25.72 & 0.268 ± 0.357 & 0.196 ± 0.280 & 0.235 ± 0.302\Tstrut\\
% \cline{2-6}
% & 96 & 27.00 & 0.296 ± 0.358 & 0.215 ± 0.285 & 0.264 ± 0.304\Tstrut\\
% \cline{2-6}
% & 128 & 27.75 & 0.332 ± 0.352 & 0.231 ± 0.282 & 0.299 ± 0.294\Tstrut\\
% \hline
% \end{tabular}
% % \end{center}
% \end{table*}

%----------------------------------

\section{Conclusions}
\label{sec:conclusion}
In this work, we proposed two progressive approaches that improve the neighborhood quality of textual data for deriving local explanations.
Our first method \gbased{}, which follows the momentum of utilizing generative models for neighborhood preparations, approximates the underlying neighborhood of input via a two-staged interpolation in the latent space maintained by a generative autoencoder.
Arise from the concern about the opacity of the generator powered by neural networks, we proposed an alternative -- \pbased{}, which also improves explanation quality by generating realistic and suitable samples.
It adopts the probability-based edition, which is fully transparent, following the local $n$-gram context as a substitute for the generator.
Both methods enhance the understanding of particular black-box decisions via explanations consisting of word-level and instance-level components.

Our experiments show, both qualitatively and quantitatively, the effectiveness of \gbased{}~in comparison to the other competitors. 
Its adoption of the generative model ensures the quality of constructed neighborhoods.
As for \pbased{}, dropping the generative model harms the validity of the created texts to a certain extent as demonstrated by the (counter)factuals in the qualitative evaluation.
But it still attains similar performance to \gbased{}~during the quantitative evaluation following the 3 \textit{C}s of interpretability.
Furthermore, the explanations are much more stable as a consequence of the transparent and controllable generation process according to Section~\ref{sec:stability}, where \gbased{}~appears to be modest.

In spite of the analysis of explanation quality, through the investigation of the dependency of \gbased{}~on the detailed settings of the generative model, we would also like to call for more caution in involving generators in explainability.
The transparency built upon opaque entities leaves potential risks in practice.
Minor changes, e.g., modifying the training scheme or altering the structure of the generator networks, could already challenge explanation quality.
Moreover, if taking biases widely reported on generative models~\cite{huang2020reducing, sheng2021societal, amini2019uncovering} into consideration, poor choices of the generator could damage more than just the quality of explanations.
For this reason, we believe an appropriate certification process is mandatory for the verification and selection of generator models deployed in explanation methods.

% \section*{Acknowledgments}
% The first author is supported by the State Ministry of Science and Culture of Lower Saxony, within the PhD program ``Responsible Artificial Intelligence in the Digital Society''. 
% We also thank Philip Naumann for the insightful discussions.

\end{document}